\documentclass[11pt]{article}

\usepackage{xspace}
\usepackage{enumerate}
\usepackage{amssymb}
\usepackage{graphicx}
\usepackage{booktabs}
\usepackage[table]{xcolor}
\usepackage{pifont}
\usepackage{pagecolor}
\usepackage{titlesec}
\usepackage{caption}
\usepackage{subcaption}
\usepackage[most]{tcolorbox}
\tcbuselibrary{listings, skins, breakable}

\usepackage{xcolor}
\usepackage[percent]{overpic} 
\usepackage{hyperref}
\usepackage[capitalize]{cleveref}

\usepackage{tikz}
\usepackage[numbers]{natbib}

\usepackage{caption}
\usepackage[utf8]{inputenc} 
\usepackage[T1]{fontenc}    
\usepackage{url}            
\usepackage{booktabs}       
\usepackage{makecell}
\usepackage{amsfonts}       
\usepackage{nicefrac}       
\usepackage{microtype}      
\usepackage{xcolor}         
\usepackage{multirow}
\usepackage{graphicx}
\usepackage{amsmath}

\usepackage{caption}
\usepackage{subcaption}
\usepackage{booktabs}

\definecolor{methodblue}{RGB}{225,240,255}
\definecolor{methodyellow}{RGB}{255,248,220}

\crefname{section}{Sec.}{Secs.}
\Crefname{section}{Section}{Sections}
\Crefname{table}{Table}{Tables}
\crefname{table}{Tab.}{Tabs.}

\usepackage{amsmath,amsfonts,bm}

\usepackage[margin=1in, top=1in]{geometry}
\usepackage{graphicx}
\usepackage{fancyhdr}
\usepackage{hyperref}
\usepackage{xcolor}
\usepackage{titlesec}
\usepackage{tcolorbox}
\usepackage{titling}
\tcbuselibrary{skins}

\usepackage[T1]{fontenc}
\usepackage{tgheros}
\usepackage[utf8]{inputenc}

\usepackage{amsmath,amssymb,bm}
\usepackage{newtxtext}
\usepackage{newtxmath} 

\AtBeginDocument{%

}

\providecommand{\titlefont}{\sffamily\bfseries}
\providecolor{qc_darkblue}{rgb}{0.008,0.063,0.247}
\providecolor{qc_blue}{rgb}{0.164,0.164,0.914}

\titleformat{\title}
{\titlefont\LARGE\color{qc_blue}}{}{0pt}{}

\titleformat{\section}
{\titlefont\Large\bfseries\color{qc_darkblue}}{\thesection}{1em}{}

\titlespacing*{\section}{0em}{1em}{.6em}

\newtcolorbox{titlebox}{
  enhanced,
  colback=white,
  boxrule=0pt,
  opacityback=0,
  opacityframe=0,
  width=0.95\textwidth,
  center
}

\makeatletter
\newcommand{\contactinfo}[1]{\def\@contactinfo{#1}}
\makeatother
\contactinfo{}

\fancypagestyle{titlepage}{
  \fancyhf{}
  \fancyfoot[L]{\footnotesize Qualcomm AI Research is an initiative of Qualcomm Technologies, Inc.}
  \fancyfoot[R]{\footnotesize\thepage}
  
}
\renewcommand\abstract{%
    \setlength{\parskip}{.8em}
    \par
}

\date{}

\contactinfo{}

\usepackage[T1]{fontenc}
\usepackage{colortbl}
\usepackage{amsfonts,bm}
\usepackage{array,multirow,makecell}
\usepackage{tabularx}
\newcolumntype{R}{>{\raggedleft\arraybackslash}X}
\newcolumntype{C}{>{\centering\arraybackslash}X}
\usepackage{microtype}

\newcommand{\speed}[1]{\ensuremath{#1\times}}
\usepackage{algorithm}
\usepackage[noend]{algpseudocode}
\algrenewcommand\algorithmicrequire{\textbf{Input:}}
\algrenewcommand\algorithmiccomment[1]{\hfill{\color{gray}\footnotesize$\triangleright$ #1}}

\hypersetup{hidelinks}

\begin{document}
\thispagestyle{titlepage}
\bibliographystyle{unsrtnat}

\begin{tikz}[remember picture,overlay]
  \node[anchor=north east, inner sep=0pt]
    at ([xshift=5cm,yshift=5.5cm]current page.north east)
    {\includegraphics[width=10cm,angle=-75,origin=c]{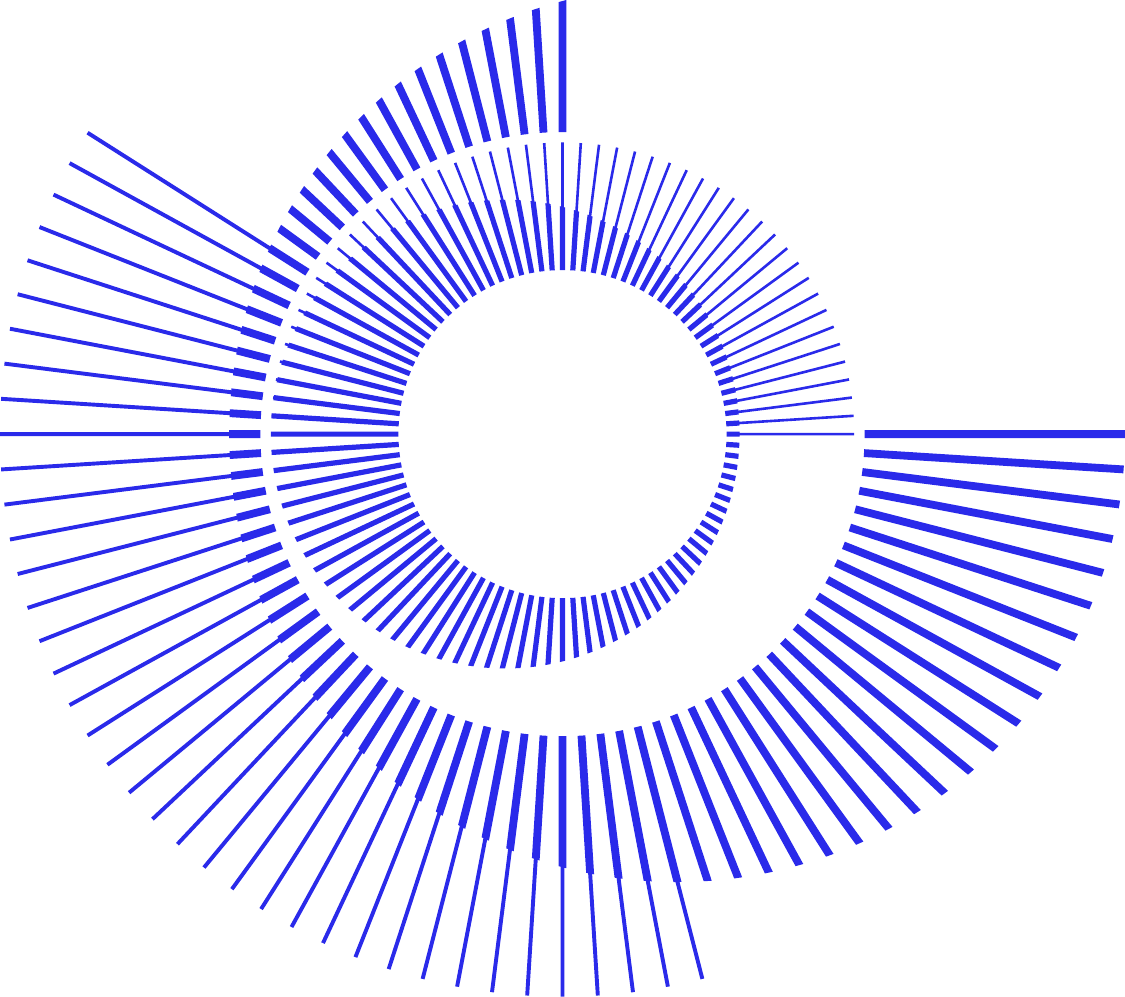}};
\end{tikz}

\begin{figure}[t]
    \vspace*{-1cm}
    \hspace*{-0.6cm} 
    \includegraphics[width=4.0cm]{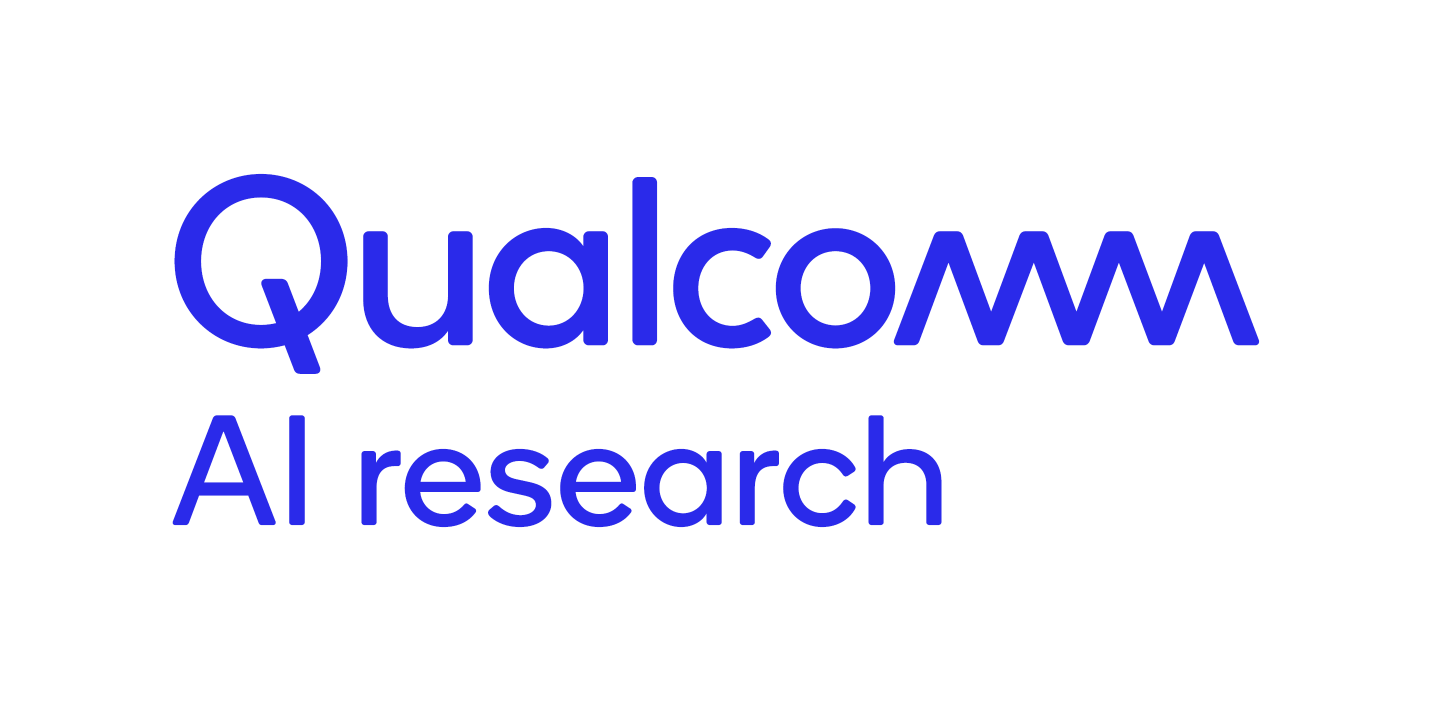}
    \vspace*{-0.5cm}
\end{figure}
\vspace{-2em}

\title{Where Compute Matters: Heterogeneous Attention for Efficient Video Diffusion}

\author{
Olga Zatsarynna$^{1,2}$,
Denis Korzhenkov$^{1}$,
Juergen Gall$^{2}$,
Amir Habibian$^{1}$,
Mohsen Ghafoorian$^{1}$
}

\contactinfo{}

\begin{titlebox}

{\titlefont\huge\bfseries\color{qc_darkblue}\thetitle}\\

\makeatletter

{\titlefont\mdseries\normalfont\color{qc_blue}
\@author
\par}

\vspace{0.4em}

{\small\normalfont\color{qc_darkblue}
$^{1}$Qualcomm AI Research
\qquad
$^{2}$University of Bonn
\par}

\ifx\@contactinfo\@empty
\else
  \vspace{-.0em}
  {\bfseries\emphfont\@contactinfo}%
\fi

\makeatother

\begin{abstract}
Efficient video generation requires reducing the quadratic cost of self-attention over long spatiotemporal token sequences. Denoising difficulty varies across video regions and evolves throughout generation, motivating an adaptive allocation of global context. We introduce HetA-DiT, a heterogeneous attention mechanism that routes uncertain tokens through global attention and processes more reliable tokens with local attention. A lightweight uncertainty branch is trained with an uncertainty-aware distribution-matching objective. At inference, its preceding-step predictions are shared across attention heads and reused across blocks, without an additional Transformer evaluation. We evaluate DMD-distilled Wan2.2-5B, Wan2.1-1.3B and Wan2.1-14B models. VBench across all three models, together with VBench2.0 and human preference evaluations on 5B, shows that HetA achieves a favorable quality-efficiency trade-off while routing only approximately 20\% of tokens through global attention. Across Wan2.1-1.3B and Wan2.2-5B/14B benchmarks on A100, HetA achieves up to \speed{3.78} self-attention and \speed{2.23} Transformer-block speedups over dense FlashAttention. Complete-denoising speedups grow with model size and sampling schedule, with composed runtime measurements projecting up to \speed{2.06} for a 14B expert with 50-step sampling.

\end{abstract}

\end{titlebox}

\section{Introduction}
\label{sec:intro}

Diffusion Transformers (DiTs) have become the dominant backbone for
high-quality video generation, powering recent open systems such as
Wan~\cite{wan2025}, HunyuanVideo~\cite{hunyuanteam2024hunyuanvideo},
CogVideoX~\cite{yang2025cogvideox}, LTX-Video~\cite{hacohen2024ltx}, and
Open-Sora Plan~\cite{lin2024opensora}. By representing a video as a long
sequence of latent spatio-temporal tokens, DiTs flexibly model appearance,
motion, and temporal consistency in a single architecture. This flexibility,
however, comes at a steep computational price: self-attention scales
quadratically with the number of tokens, and video sequences are typically an
order of magnitude longer than image sequences. As a result, a single
denoising step is dominated by attention cost. The community has responded
with a growing set of efficiency techniques, many of which make
attention cheaper \emph{uniformly}, treating every token as equally important
to compute.

\begin{figure}[t!]
    \centering
    \includegraphics[width=0.6\linewidth]{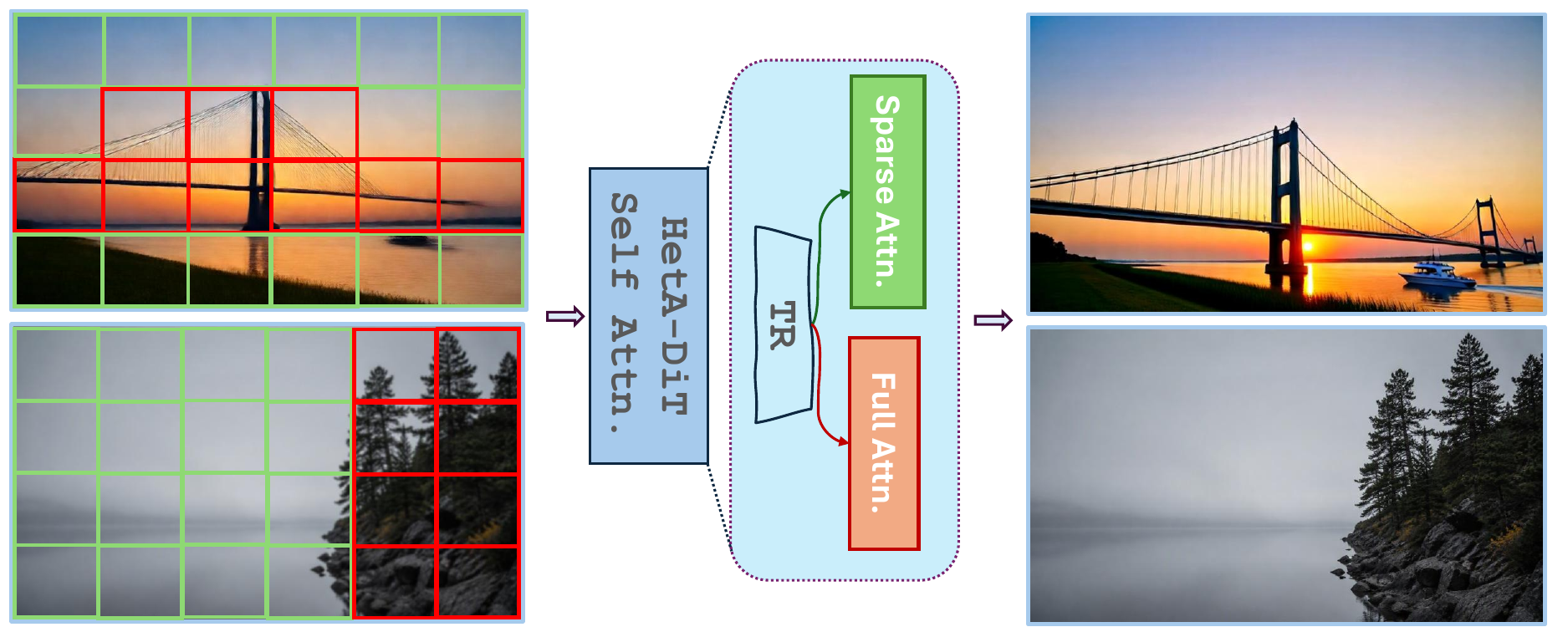}
    \caption{\textbf{Spend compute where it counts.} Not all video tokens are
    equally hard to denoise; fine structures demand global context, while
    smooth regions are handled by cheap local attention. Our heterogeneous
    attention method incorporates this via an adaptive uncertainty-based token
    router (TR) that dispatches each token to local or dense attention.}
    \label{fig:teaser}
\end{figure}

A key observation behind our work is that this uniformity is at odds with the
structure of video content: \emph{not all spatio-temporal tokens are equally
hard to denoise}. Large regions such as static backgrounds, sky, or smoothly
varying textures can be predicted reliably from local context very early in
denoising. Other regions---moving objects, occlusions, object boundaries,
faces, hands, and fine semantic details---remain ambiguous much longer and
benefit from richer global context. Equally important, \emph{which}
tokens are hard is not fixed: it changes across denoising steps, across
prompts, and across frames as motion and occlusion evolve. Any static sparsity
pattern is therefore fundamentally mismatched to the underlying problem.
Figure~\ref{fig:teaser} illustrates this behavior.

Beyond making attention uniformly cheaper, a less explored direction is to allocate compute \emph{non-uniformly} across the tokens themselves. Token caching~\citep{zou2024toca,shmilovich2025liteattention} skips stable tokens but relies on redundancy across many denoising steps, which diminishes under step distillation. Feed-forward mixture-of-experts DiTs~\citep{fei2024dimoe} route the FFN sub-layer, leaving attention unchanged. Training-free adaptive sparsification~\citep{peruzzo2025adaptor,liu2025astraea,zhang2025spargeattn} avoids tuning, but the model is not adapted to its sparsity pattern. All-or-Here Attention (AHA)~\citep{luo2025aha} learns token/head gates between local and global attention in language models, recomputing them at every layer. We target the design point between these: \emph{adaptive, attention-level, DMD-compatible} token routing, obtained with a small adaptation cost of approximately 12 H100 GPU-days.

We propose a \textbf{heterogeneous attention} mechanism that performs
non-uniform token processing. We augment the denoiser with a lightweight
uncertainty prediction branch that produces a per-token scalar, trained with
an uncertainty-aware reconstruction loss so that high values mark tokens whose
clean-latent prediction is unreliable. After training, the branch is frozen
and used only to produce a stable routing signal. At inference, an adaptive
per-sample, per-timestep threshold sends high-uncertainty tokens to dense
self-attention and reliable tokens to local self-attention, with a single
hyperparameter trading efficiency for quality. Three properties distinguish
this design: (i) routing is \emph{content- and timestep-adaptive}, so dense
compute follows difficult regions across frames and denoising steps; (ii) routing is \emph{inexpensive}, using preceding-step predictions shared across heads and reused across blocks without an additional Transformer evaluation;
and (iii) it is explicitly \emph{compatible with step distillation} and is
demonstrated on DMD-distilled Wan models.

Applied to DMD-distilled Wan2.2-5B at 704p and Wan2.1-1.3B and Wan2.1-14B at 480p~\citep{wan2025}, HetA keeps approximately 80\% of queries local. On 5B, it scores 83.28 versus 82.94 for dense three-step DMD on VBench~\citep{huang2023vbench} and improves over dense two-step DMD on both VBench and VBench2.0~\citep{zheng2025vbench}. In operator benchmarks, our implementation makes self-attention $2.05\times$ and the Transformer block $1.30\times$ faster than dense FlashAttention on 5B, rising to \textbf{$3.78\times$} and \textbf{$2.23\times$} on a Wan2.2-14B expert at 704p, with composed-runtime projections of up to $2.06\times$ for 50-step sampling. On Wan2.1-14B, attention-LoRA adaptation reaches 84.01 on full VBench versus 84.63 for dense DMD3, while improving the semantic score from 81.97 to 82.29. On 5B, HetA is competitive on VBench with efficient-attention and caching baselines, including VSA, Jenga, ToCa, Attention Surgery and ReHyAt, and evaluators prefer it over Attention Surgery in 68\% of pairs versus 19\%. Matched ablations against uniform routing and adapted AHA attribute the gain to the uncertainty signal.

Our contributions are:
\begin{itemize}
\item \textbf{HetA-DiT}, an uncertainty-routed heterogeneous attention mechanism: uncertain tokens attend globally and the rest locally, with a single threshold trading quality for efficiency. Routes are shared across heads and reused across blocks and steps, adding no Transformer evaluation and remaining compatible with few-step DMD.
\item An \textbf{uncertainty-aware DMD objective} that trains a lightweight per-token uncertainty branch without paired ground-truth videos, validated directly by a replicated global-context sensitivity test.
\item \textbf{Extensive evaluation} with full VBench on Wan2.2-5B, Wan2.1-1.3B and Wan2.1-14B, plus VBench2.0 and a blinded human study on 5B. On 5B, HetA performs competitively on VBench with other efficient methods and with dense DMD, while approximately 80\% of queries are routed to the local path; matched ablations attribute its quality to uncertainty routing.
\item An \textbf{optimized Triton implementation} with up to $3.78\times$ attention and $2.23\times$ block speedups over dense FlashAttention, and end-to-end speedups that grow with model size and sampling schedule.
\end{itemize}

\section{Related Work}
\paragraph{Efficient Video Diffusion.}
Open-source video diffusion models such as
Wan~\cite{wan2025}, HunyuanVideo~\cite{hunyuanteam2024hunyuanvideo},
CogVideoX~\cite{yang2025cogvideox}, LTX-Video~\cite{hacohen2024ltx}, and Open-Sora Plan~\cite{lin2024opensora} have driven rapid progress in text-to-video generation, but their DiT backbones incur a large computational
cost that grows quadratically with the number of spatio-temporal tokens. A first line of efficiency work reduces the \emph{number} of denoising steps via step distillation: DMD~\cite{yin_one-step_2024,yin_improved_2024,korzhenkov2026pyramidalwan} and follow-ups~\cite{pmlr-v267-lin25m,fan2026phaseddmdfewstepdistribution,nie2026transitionmatchingdistillationfast,zheng2026large} now yield $>$30$\times$ speedups and have become the de facto deployment regime.
A parallel line targets deployment on edge and mobile hardware through step distillation, sparse/hybrid attention, model pruning and quantization, e.g., Neodragon~\cite{karnewar2026neodragon}, SnapGen-V~\cite{wu2025snapgen}, MobileVD~\cite{Ben_Yahia_2025_ICCV}, AMD-Hummingbird~\cite{isobe2025amd}, Video DiT~\cite{wu2025taming}, and MobileWan~\cite{ghafoorian2026mobilewan}. All of these methods, however, treat every token identically within a denoising step.

\paragraph{Efficient Attention.}
A large body of work reduces the cost of self-attention within a step through content-agnostic mechanisms. Fixed sparse patterns date back to classical efficient transformers~\cite{child2019sparse,beltagy2020longformer,
zaheer2020bigbird} and have been adapted to video DiTs through Sliding Tile Attention~\cite{zhang2025fast}. Trainable-sparse variants such as VSA~\cite{zhang2025faster} and SLA~\cite{zhang2025sla} learn a sparse mask,
while linear-attention backbones~\cite{katharopoulos2020transformers, wang2020linformer,choromanski2020performer,xiong2021nystromformer} have recently been extended to video with SANA-Video~\cite{chen2025sana} and
HLA~\cite{ackermann2026hla}. Hybrid recurrent designs such as ReHyAt~\cite{ghafoorian2026rehyat} and Attention Surgery~\cite{ghafoorian2026attentionsurgery}, and architectural swaps like Mamba-based  M4V~\cite{huang2025m4v}, follow the same philosophy of making attention globally cheaper. These methods mainly reduce the cost of attention using a common execution policy, rather than allocating a local or global attention operator according to predicted denoising uncertainty.

\paragraph{Caching.}
Feature caching methods exploit temporal redundancy across denoising steps. ToCa~\cite{zou2024toca} caches activations for tokens deemed stable and recomputes the rest; LiteAttention~\cite{shmilovich2025liteattention} does the same at the level of attention; Clockwork Diffusion~\cite{habibian2024clockwork} and object-centric editing~\cite{kahatapitiya2024object} reuse coarse features across timesteps. Two properties distinguish caching from our method. First, caching makes a \emph{hard binary} cache-vs-compute decision, whereas we allocate a \emph{soft} dense-vs-local budget that still updates every token. Second, caching amortizes skipped compute over many denoising steps; in the low-step regime this redundancy largely vanishes, reducing the opportunities for caching. DisCa~\citep{zou2026disca} studies learnable caching compatible with distillation; HetA instead updates every token using current local or global context.

\paragraph{Mixture of Experts.}
A common design in mixture-of-experts DiTs~\cite{fei2024dimoe} routes tokens through \emph{feed-forward} experts, leaving self-attention unchanged. This token-wise allocation is conceptually related to our routing. In video DiTs, self-attention is the bottleneck because of quadratic scaling in the token count. For example, in Wan2.2 A14B at 720p resolution, 82\% of the compute is spent on self-attention while only 13\% goes to the FFN. FFN-level MoE therefore leaves the dominant attention cost untouched. Our method is complementary: it routes at the \emph{attention}
sub-layer and could, in principle, be combined with FFN-level MoE.

\paragraph{Adaptive attention allocation.}
All-or-Here Attention (AHA)~\citep{luo2025aha} learns token/head gates between local and global attention for language models. HetA uses uncertainty supervised at the whole-denoiser output and shares preceding-step routes across heads and blocks. Our adapted AHA comparison matches the video neighborhood, initialization, random swaps and training budget, isolating the complete routing designs. Efficient execution also matters: dense FlashAttention~\citep{dao2022flashattention} avoids materializing attention matrices, while HetA separates local and global query execution to turn a lower global-query budget into GPU savings.

\section{Method}
\label{sec:method}

In this work, we propose allocating compute \textit{where it matters} on a per-token basis. Generated videos suggest unequal denoising complexity: some regions are smooth and uniform, while others contain complex patterns and textures. Motivated by this observation, we allocate computational budget according to token complexity, focusing on the attention operation, one of the main bottlenecks of DiTs.

To this end, we equip a few-step DMD model with an additional uncertainty branch that serves as a proxy for complexity prediction. We then use this branch to route tokens to either sparse local attention or full attention within the underlying DiT model, which we then term HetA (Heterogeneous self Attention) DiT. This preserves global context for difficult regions while reducing unnecessary dense computation on easier tokens. We first briefly review self-attention and DMD training for video diffusion models, and then describe our proposed uncertainty-based heterogeneous HetA DiT model.

\subsection{Preliminaries.}

\subsubsection{Self-Attention.}
\label{sec:prelim_attention}

The key operation in video DiT models is self-attention, which relates all tokens to one another. Formally, given hidden tokens $Z \in \mathbb{R}^{N \times d}$, a transformer block computes queries, keys, and values:
\begin{equation}
Q = ZW_Q,\qquad K = ZW_K,\qquad V = ZW_V.
\end{equation}
Self-attention then updates token $i$ using all tokens:
\begin{equation}
\mathrm{Attn}_{\mathrm{dense}}(i)
= \sum_{j=1}^{N} a_{ij} v_j,
\ \ \ \ \ \ \ \ a_{ij} =
\frac{
\exp(q_i^\top k_j / \sqrt{d_h})
}{
\sum_{\ell=1}^{N} \exp(q_i^\top k_\ell / \sqrt{d_h}),
}
\label{eq:dense_attention}
\end{equation}
where $d_h$ is the channel dimension.
This operation is expressive but has quadratic cost in the number of video tokens, $\mathcal{O}(N^2 d_h)$, making it expensive for long or high-resolution videos.
\textbf{Local} self-attention, on the other hand, restricts the key-value set to a spatio-temporal neighborhood $\mathfrak{N}(i)$ around token $i$:
\begin{equation}
\mathrm{Attn}_{\mathrm{local}}(i)
=
\sum_{j \in \mathfrak{N}(i)} \tilde{a}_{ij} v_j.
\label{eq:local_attention}
\end{equation}

If $|\mathfrak{N}(i)|=M \ll N$, local attention has cost $\mathcal{O}(NMd_h)$. However, purely local attention may lose global information that is important for object identity, occlusions, fast motion, and temporal consistency. Our goal is therefore to use dense attention only where it is needed.

\subsubsection{Distribution Matching Distillation.}
\label{sec:prelim_dmd}

Distribution Matching Distillation (DMD) trains a student generator to match the distribution of a pretrained diffusion teacher using fewer denoising steps~\cite{yin_one-step_2024}. Given a prompt $y$ and noise $z$, the student produces a latent video
\begin{equation}
\tilde{x}_0 = G_\theta(z,y).
\end{equation}
The sample is then re-noised to timestep $\tau$:
\begin{equation}
\tilde{x}_\tau = \alpha_\tau \tilde{x}_0 + \sigma_\tau \epsilon,
\qquad
\epsilon \sim \mathcal{N}(0,I),
\end{equation}
where $\alpha_\tau$ and $\sigma_\tau$ define the teacher diffusion schedule. The teacher score $s_{\mathrm{T}}(\tilde{x}_\tau,\tau,y)$ provides the direction that moves the student sample toward the teacher distribution. DMD optimizes the student using a distribution-matching gradient of the form
\begin{equation}
\nabla_\theta \mathcal{L}_{\mathrm{DMD}}
=
\mathbb{E}_{z,y,\tau,\epsilon}
\Big[
w(\tau)
\Big(
s_{\mathrm{S}}(\tilde{x}_\tau,\tau,y)-
s_{\mathrm{T}}(\tilde{x}_\tau,\tau,y)
\Big)
\nabla_\theta  \tilde{x}_0
\Big],
\label{eq:regular_dmd_loss}
\end{equation}
where $s_{\mathrm{S}}$ is the student score and $w(\tau)$ is a timestep-dependent weight. Thus, the student is updated so that its generated video distribution matches the teacher distribution. For video generation, DMD is applied in the video latent space, and supervision comes from the pretrained teacher rather than from paired ground-truth videos.

Although DMD fine-tuning already drastically reduces the number of denoising steps required to obtain high-quality generated videos, we further optimize the underlying DiT model to enable even more efficient generation.

\subsection{Proposed Approach.}
\label{sec:proposed_approach}

Overall, our proposed approach consists of two stages. First, we train a full DiT model augmented with an uncertainty branch using a modified DMD objective. Then, we rely on this branch to introduce heterogeneous attention computation into the underlying DiT - resulting in our final proposed HetA-DiT model. We describe both stages in detail below.

\subsubsection{Stage 1: Uncertainty Prediction.}
\label{sec:uncertainty_branch}
Our goal is to differentiate between tokens that are easier or more difficult to denoise. To this end, we extend the transformer with an additional branch that estimates the complexity of each latent token (after patchification). Formally, for each noisy latent $x_\tau$, the model predicts both a denoised latent $\tilde{x}_0$ and a token-wise uncertainty map $u$:
\begin{equation}
(\tilde{x}_0,u) = f_\theta(x_\tau,\tau,y),
\quad
u = \{u_i\}_{i=1}^{N},
\quad
u_i \geq 0.
\label{eq:uncertainty_prediction}
\end{equation}
The scalar $u_i$ estimates the denoising uncertainty of token $i$.

To integrate uncertainty prediction with the DMD gradient in Eq.~\ref{eq:regular_dmd_loss}, we define a stopped-gradient pseudo-target,
\begin{equation}
\hat{x}_0=\operatorname{stopgrad}\!\left[
\tilde{x}_0-w(\tau)\bigl(s_{\mathrm{S}}(\tilde{x}_\tau,\tau,y)
-s_{\mathrm{T}}(\tilde{x}_\tau,\tau,y)\bigr)\right].
\label{eq:pseudo_target}
\end{equation}
The DMD gradient can then be written as a reconstruction gradient toward this pseudo-target:
\begin{equation}
    \nabla_\theta \mathcal{L}_{\mathrm{DMD}} = \mathbb{E}_{z,y,\tau,\epsilon} \left[
    \nabla_\theta \frac{1}{2}\left\lVert \tilde{x}_0 - \hat{x}_0 \right\rVert^2
    \right].
\label{eq:mse_dmd_loss}
\end{equation}
This formulation can be naturally generalized to per-token negative Gaussian log-likelihood with predicted variance,
\begin{equation}
    \mathcal{L}_{\mathrm{DMD}}^{\mathrm{unc}} = \mathbb{E}_{z,y,\tau,\epsilon} \left[
    - \sum_{i=1}^N \log \mathcal{N}\left(\hat{x}_{0, i} \mid \tilde{x}_{0, i}, u_i^2\right)
    \right].
\end{equation}
As commonly known, Gaussian log-likelihood consists of the two non-constant terms, 
\begin{equation}
    - \log \mathcal{N} \left(\hat{x}_{0, i} \mid \tilde{x}_{0, i}, u_i^2\right) = 
    \ \lambda_{\textrm{rec}} \frac{
        \|\tilde{x}_{0,i}-\hat{x}_{0,i}\|^2
        }{
        u_i^2
        }
    +
    \lambda_{\log}\log u_i
    + \mathrm{const}.
\end{equation}

\begin{figure*}[t!]
\centering
\includegraphics[width=0.80\linewidth]{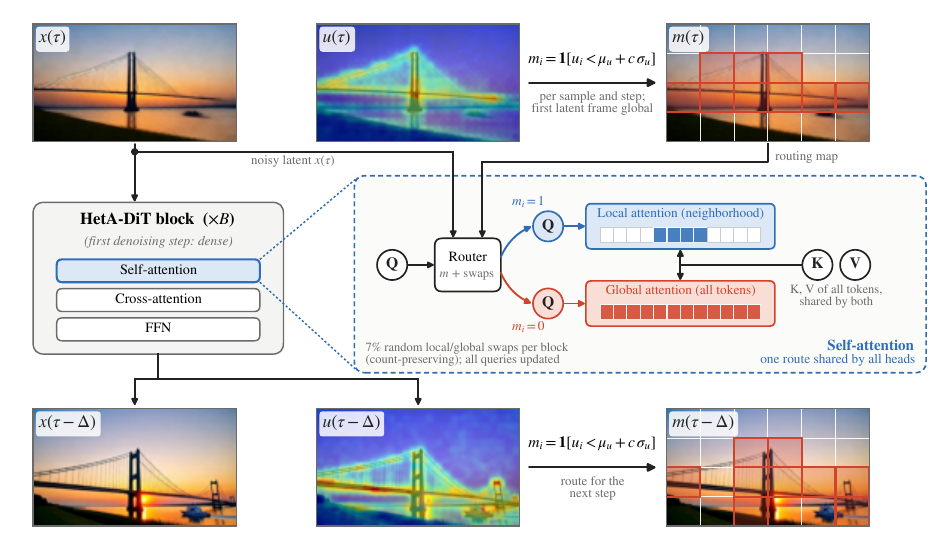}
\caption{\textbf{HetA-DiT.} Our approach assigns heterogeneous self-attention computation to tokens with different denoising complexities. Given the uncertainty map $u(\tau)$ predicted at step $\tau$ by the dedicated branch, we construct a routing mask $m(\tau)$ by thresholding the predicted uncertainties. Based on this mask $m(\tau)$ and the latent $x(\tau)$, tokens with low uncertainty are processed using cheaper local self-attention in the following denoising step $\tau-\Delta$, while high-uncertainty tokens are processed with full attention. The resulting outputs are then used to determine heterogeneous self-attention processing in the next denoising step.}
\label{fig:main}
\end{figure*}

The first term is proportional to the former MSE objective~(\ref{eq:mse_dmd_loss}) and encourages larger uncertainty where the prediction error is high, while the second term prevents the trivial solution of predicting arbitrarily large uncertainty. 
Notably, we found that with this uncertainty-based weighting we do not need the per-sample DMD loss weighting which is otherwise commonly used in practice~\cite{yin_one-step_2024}.

\subsubsection{Stage 2: Uncertainty-Guided Token Densification.}
\label{sec:token_densification}

After pretraining the uncertainty branch, we freeze it and use its predictions during the second DiT fine-tuning stage to construct a routing mask that assigns heterogeneous attention patterns to tokens, as illustrated in Figure~\ref{fig:main}.

Specifically, given a timestep $\tau$, the branch predicts token-wise uncertainties $\{u_i(\tau)\}_{i=1}^{N}$. We compute their mean $\mu_u(\tau)$ and standard deviation $\sigma_u(\tau)$, and define a per-step threshold for heterogeneous attention assignment:
\begin{equation}
T(\tau)
=
\mu_u(\tau) + c\,\sigma_u(\tau).
\label{eq:threshold}
\end{equation}
Here, $c$ is a hyperparameter controlling the number of tokens assigned to full attention. Smaller values of $c$ lower the threshold and route more tokens to full attention, whereas larger values make full-attention assignment less likely. Based on this threshold, we define a binary mask:
\begin{equation}
m_i(\tau)
=
\mathbf{1}
\left[
u_i(\tau) < T(\tau)
\right].
\label{eq:mask}
\end{equation}
Given this mask, uncertain tokens with $m_i=0$ are processed with full attention, while certain tokens with $m_i=1$ are processed locally.
Furthermore, to increase token-wise coverage of full attention, at each transformer block $b$, we additionally select a subset of uncertain tokens and swap them with randomly selected certain tokens. This exposes more tokens to global processing over the course of generation, while ensuring that uncertain tokens still receive full attention most of the time.

Finally, the predicted mask is used for the next denoising step $\tau-\Delta$ to route the heterogeneous self-attention computation, where the resulting self-attention update can be written as:
\begin{equation}
z_i'
=
m_i\mathrm{Attn}_{\mathrm{local}}(i)
+
(1-m_i)\mathrm{Attn}_{\mathrm{dense}}(i).
\label{eq:final_update}
\end{equation}
We fine-tune the resulting sparsified model using the standard DMD objective, as described in Sec.~\ref{sec:prelim_dmd}.

In terms of computational savings, let $\rho = \frac{1}{N}\sum_i (1-m_i)$ be the fraction of dense-query tokens, and let $M=|\mathfrak{N}(i)|$ be the local window size. The resulting attention cost then can be estimated as:
\begin{equation}
\mathcal{O}
\left(
\rho N^2 d_h + (1-\rho)NMd_h
\right),
\label{eq:cost}
\end{equation}
compared with $\mathcal{O}(N^2d_h)$ for fully dense attention. Thus, when only a small fraction of tokens is uncertain and $M \ll N$, the model reduces attention cost while preserving dense global context for the tokens that actually require it. 
In our experiments the dense ratio $\rho \approx 0.2$.

Routes are shared across attention heads and reused across blocks; each block applies fresh random local/global exchanges. The first denoising step uses dense attention, and the first latent frame remains global. The output-change diagnostic in Section~\ref{sec:uncertainty_results} evaluates the preceding-step signal used by this routing policy.

\begin{algorithm}[t]
\caption{HetA-DiT few-step inference}
\label{alg:heta}
\small
\begin{algorithmic}[1]
\Require prompt $y$, noise $x_{\tau_1}$, timesteps $\tau_1>\dots>\tau_S$, threshold $c$, swap rate $s$
\State $(\tilde x_0,u)\gets f_\theta(x_{\tau_1},\tau_1,y)$ \Comment{dense first step}
\For{$k=2,\dots,S$}
  \State $x_{\tau_k}\gets$ Euler step from $x_{\tau_{k-1}}$ using $\tilde x_0$ \Comment{flow matching}
  \State $m\gets\mathbf 1[u<\mu_u+c\,\sigma_u]$ on frames${}\ge2$ \Comment{frame 1 global}
  \For{each DiT block}
    \State $m'\gets$ swap $sN$ local/global pairs of $m$
    \State $z_{\{m'_i=1\}}\gets\mathrm{Attn}_{\mathrm{local}}(Q_{\{m'_i=1\}},K,V)$ \Comment{$\mathfrak N(i)$}
    \State $z_{\{m'_i=0\}}\gets\mathrm{Attn}_{\mathrm{full}}(Q_{\{m'_i=0\}},K,V)$ \Comment{all $K,V$}
    \State scatter $z$; apply cross-attention and FFN
  \EndFor
  \State $(\tilde x_0,u)\gets$ outputs of this pass \Comment{$u$ routes step $k{+}1$}
\EndFor
\State \Return $\tilde x_0$
\end{algorithmic}
\end{algorithm}

\subsection{Efficient GPU Execution}
\label{sec:kernel}
We implement local and global attention as separate GPU calls. A Triton kernel processes compact local-query tiles and their 3D neighborhoods. Global queries are gathered into one list shared across heads and processed with FlashAttention against all original keys and values; both outputs are scattered to their original positions. Separating the calls matters because global queries are scattered across the video: on saved routes from an uncertainty-guided generation, 99.996\% of flat 64-query tiles, and 91.9\% of spatial 16-query tiles, would contain at least one global query in a mixed layout, forcing nearly every tile to load all keys. Separate calls keep the local tiles purely local. Neighborhood geometry is cached, while query assignments are rebuilt after each block's swaps. The kernel preserves the attention rule and is checked against an independent reference; detailed tile sizes and timing protocols are in the appendix.

\section{Experimental Setup}
\label{sec:experimental_setup}

\begin{figure*}[t!]
    \centering
    \includegraphics[width=0.9\linewidth]{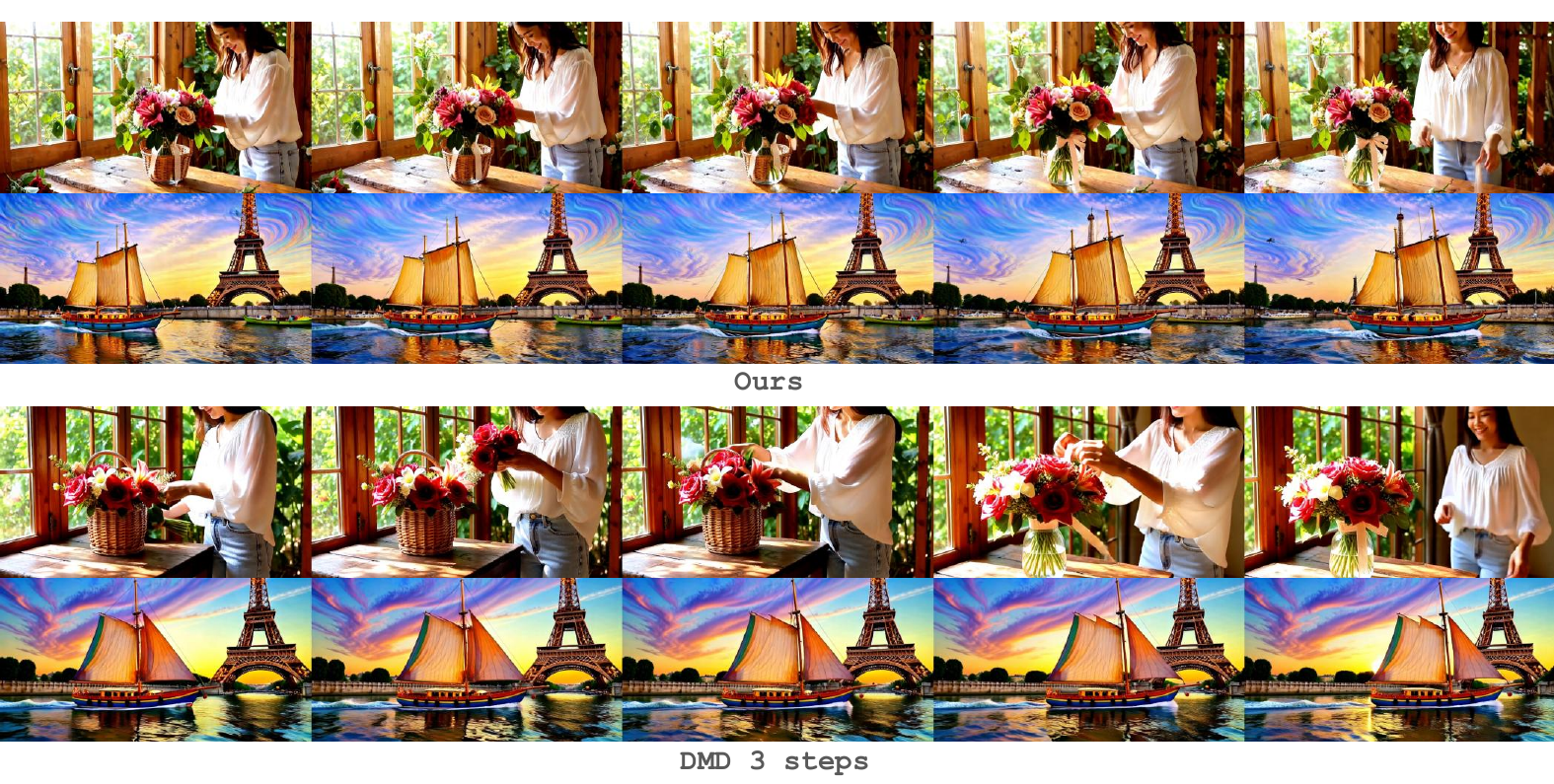}
    \caption{\textbf{Qualitative comparison of HetA-DiT to baseline DMD}. Both models run 3-step denoising. Our method maintains the high quality of the full step-distilled model while being substantially more efficient in an already low-compute regime.}
    \label{fig:qual}
\end{figure*}

\paragraph{Training Setting. }
We train our proposed method in two stages. First, we train a DMD model augmented with an additional uncertainty branch using the modified DMD objective described in Sec.~\ref{sec:uncertainty_branch}. The first-stage training is performed for 15k alternating iterations (7,500 generator updates). 
In the second stage, we freeze the pre-trained uncertainty branch and fine-tune the remaining DiT parameters for 6k alternating iterations (3,000 generator updates) using the standard DMD objective, with heterogeneous sparsification guided by the branch predictions. Since the routing mask at each denoising step is based on uncertainty from the previous step, training uses an additional full-attention student forward pass without gradient propagation to estimate it. Specifically, for a randomly sampled timestep, we predict the uncertainty at an earlier timestep and use it to construct the sparse attention mask for the current step. At inference time, no extra transformer pass is needed, as uncertainty predictions from the previous denoising step are reused to build the mask for the next step.

For both stages, we use 4 H100 GPUs and the original pretrained Wan model as the teacher. We also follow a prompt-only training protocol: visual supervision is provided by the pretrained generative model rather than ground-truth videos. This allows us to train the model without requiring direct access to video annotations. For prompts, we generate captions for a subset of videos from the VIPE1M dataset~\cite{huang2025vipe} using the Qwen3 model~\cite{yang2025qwen3}. We provide further training details in the appendix. For Wan2.1-14B, the uncertainty stage trains its prediction heads for 500 generator updates, followed by 3,000 updates to the existing attention LoRA parameters with uncertainty frozen. The adaptation uses training noise shift 5; both stages retain the dense LoRA-distilled initialization.

\paragraph{Evaluation Setting.}
To evaluate our proposed approach, we fine-tune HetA-DiT starting from DMD models based on Wan2.2-5B, Wan2.1-1.3B and Wan2.1-14B~\cite{wan2025}. For comparison with other efficient diffusion variants, we fine-tune training-based methods under DMD step distillation and evaluate training-free approaches in the same few-step setting. Unless specified otherwise, efficient variants use three denoising steps and the original Wan decoder. We generate $121\times704\times1280$ videos for 5B and $81\times480\times832$ for 1.3B and 14B. All three models are evaluated on full VBench~\cite{huang2023vbench}; the VBench2.0~\cite{zheng2025vbench} and human comparisons evaluate the 5B models.

To complement the automatic metrics, we conduct a \textbf{method-blinded human preference study}. Evaluators were presented with GPT-enhanced long prompts from VBench and asked to select the preferred video, or mark the comparison as a tie, based on semantic coherence and visual quality. For each prompt, videos from the two methods were presented side by side, with their left-right placement randomized independently across questions. In total, the study comprised 1,248 paired comparisons from 19 distinct evaluators, covering 1,082 unique videos and 667 unique prompts. The numbers of videos and prompts differ because some prompts were evaluated with multiple random seeds.

\paragraph{Compute Complexity Assessment.}
We analyze how the proposed heterogeneous attention mechanism reduces the number of \textbf{floating-point operations} compared to the full DiT model, both at the attention level and at the full Transformer block level. We also compare the resulting speedup against other efficient attention methods. We utilize DeepSpeed library to measure the total floating point operations.

In addition to FLOPs reductions, we also show that our method reduces the actual \textbf{latency} of the Transformer. We use separate Triton local-attention and FlashAttention global-attention calls, and compare against dense FlashAttention. Operator timings include live routing, swaps, gathering and scattering. All reported GPU latency measurements use A100-SXM4-80GB, BF16 and batch size one; the appendix describes the operator and complete-generation protocols. 

\paragraph{Matched routing and timing.}
The uniform and adapted AHA controls use the same dense initialization, $11^3$ neighborhood, 7\% swaps and 3,000-generator-update adaptation budget as HetA. AHA preserves head-specific gates and selects its top-scoring 20\% eligible positions; HetA's threshold selects approximately 18\% before swaps. Quality comparisons therefore test the complete routing designs.

Full VBench uses 944 GPT-enhanced prompts, five samples per prompt and all 16 dimensions. For the 5B routing and AHA comparisons, intervals use 20,000 paired prompt resamples, keeping each prompt's five samples together and recomputing the official normalized and weighted total. These intervals capture prompt variation for one trained model per method.

\begin{table*}[t]
\centering
\begin{minipage}[t]{0.49\textwidth}
\vspace{0pt}
\centering\footnotesize
\setlength{\tabcolsep}{3pt}
\begin{tabularx}{\linewidth}{@{}lrRR@{}}
\toprule
Model / efficient variant & Total$\uparrow$ & Quality$\uparrow$ & Semantic$\uparrow$ \\
\midrule
\multicolumn{4}{l}{Models with 2B-5B parameters} \\
Open-Sora Plan V1.3~\citep{lin2024opensora} & 77.23 & 80.14 & 65.62 \\
Open-Sora V1.2~\citep{zheng2024open} & 79.76 & 81.35 & 73.39 \\
LTX-Video~\citep{hacohen2024ltx} & 80.00 & 82.30 & 70.79 \\
SnapGenV~\citep{wu2025snapgen} & 81.14 & 83.47 & 71.84 \\
Hummingbird~\citep{isobe2025amd} & 81.35 & 83.73 & 71.84 \\
Mobile Video DiT (mobile)~\citep{wu2025taming} & 81.45 & 83.12 & 74.76 \\
CogVideoX 2B~\citep{yang2025cogvideox} & 81.55 & 82.48 & 77.81 \\
Neodragon~\citep{karnewar2026neodragon} & 81.61 & 83.68 & 73.36 \\
PyramidalFlow~\citep{jin2024pyramidalflow} & 81.72 & 84.74 & 69.62 \\
CogVideoX 5B~\citep{yang2025cogvideox} & 81.91 & 83.05 & 77.33 \\
CogVideoX1.5 5B~\citep{yang2025cogvideox} & 82.01 & 82.72 & 79.17 \\
Mobile Video DiT (server)~\citep{wu2025taming} & 83.09 & 84.65 & 76.86 \\
Wan2.2-5B (50 steps)~\citep{wan2025} & 83.28 & 85.03 & 76.28 \\
\midrule
\multicolumn{4}{l}{Efficient variants at $704\times1280$} \\
STA (HunyuanVideo)~\citep{zhang2025fast} & \underline{83.00} & \textbf{85.37} & 73.52 \\
Dense DMD2 (2 steps) & 82.48 & 83.05 & 80.22 \\
Dense DMD3 (3 steps) & 82.94 & 83.56 & 80.43 \\
Attention Surgery~\citep{ghafoorian2026attentionsurgery} & 82.62 & 83.60 & 78.71 \\
ReHyAt~\citep{ghafoorian2026rehyat} & 82.55 & 82.78 & \textbf{81.61} \\
VSA~\citep{zhang2026faster} & 82.24 & 82.71 & 80.38 \\
\rowcolor{methodyellow}ToCa~\citep{zou2024toca} & 80.52 & 81.40 & 76.98 \\
\rowcolor{methodyellow}Jenga~\citep{zhang2026training} & 82.72 & 83.58 & 79.22 \\
AHA (adapted)~\citep{luo2025aha} & 82.56 & 83.02 & 80.70 \\
\rowcolor{methodblue}HetA-DiT & \textbf{83.28} & \underline{83.71} & \underline{81.52} \\
\bottomrule
\end{tabularx}
\caption{\textbf{Full VBench comparison with video models and efficient variants.} Literature rows use their reported settings. Reproduced variants use three-step Wan2.2-5B except DMD2; STA uses its reported HunyuanVideo setting. Bold/underline rank efficient variants. Yellow denotes training-free variants. Paired prompt interval for HetA vs.\ AHA: is computed as $[+0.36,+1.08]$.}
\label{tab:vbench_sota}
\end{minipage}\hfill
\begin{minipage}[t]{0.49\textwidth}
\vspace{0pt}
\centering\footnotesize
\setlength{\tabcolsep}{3pt}
\begin{tabularx}{\linewidth}{@{}lRRR@{}}
\toprule
Wan2.1-1.3B & Total$\uparrow$ & Quality$\uparrow$ & Semantic$\uparrow$ \\
\midrule
Wan (reported) & 83.31 & 85.23 & 75.65 \\
Wan (reproduced) & 83.10 & 85.10 & 75.12 \\
VSA & 82.77 & 83.60 & \textbf{79.47} \\
Dense DMD3 & \textbf{83.61} & \textbf{84.82} & \underline{78.78} \\
\rowcolor{methodblue}HetA-DiT & \underline{83.10} & \underline{84.23} & 78.55 \\
\bottomrule
\end{tabularx}
\caption{\textbf{Full VBench on 1.3B.} Videos have $81\times480\times832$ frames/pixels. Bold/underline rank adapted efficient variants.}
\label{tab:vbench_sota_wan2.1}
\medskip
\begin{tabularx}{\linewidth}{@{}lCCC@{}}
\toprule
Compared baseline & \makecell{HetA\\wins} & Tie & \makecell{Baseline\\wins} \\
\midrule
Dense DMD2 & 50\% & 26\% & 24\% \\
Dense DMD3 & 32\% & 32\% & 36\% \\
Attention Surgery & 68\% & 13\% & 19\% \\
\bottomrule
\end{tabularx}
\caption{\textbf{Blinded human preference} over 1,248 video pairs. Each row shows preference against the indicated baseline.}
\label{tab:human_preference}
\medskip
\begingroup
\setlength{\tabcolsep}{2pt}
\begin{tabularx}{\linewidth}{@{}lRRRRRR@{}}
\toprule
\multirow{2}{*}{Window} & \multirow{2}{*}{Total$\uparrow$} & \multirow{2}{*}{Qual.$\uparrow$} & \multirow{2}{*}{Sem.$\uparrow$} & \multicolumn{1}{c}{Attn.} & \multicolumn{1}{c}{Attn.} & \multicolumn{1}{c}{Block} \\
& & & & \multicolumn{1}{c}{FLOPs} & \multicolumn{1}{c}{speedup} & \multicolumn{1}{c}{speedup} \\
\midrule
$13^3$ & \textbf{83.36} & \textbf{83.99} & 80.82 & $4.76\times$ & $1.80\times$ & $1.24\times$ \\
\rowcolor{methodblue}$11^3$ & 83.28 & 83.71 & \textbf{81.52} & $4.85\times$ & $2.05\times$ & $1.30\times$ \\
$9^3$ & 83.22 & 83.89 & 80.54 & $4.91\times$ & $\mathbf{2.48\times}$ & $\mathbf{1.37\times}$ \\
\bottomrule
\end{tabularx}
\endgroup
\caption{\textbf{Window-size ablation on Wan2.2-5B}, with 7\% swaps. Attention/block speedups divide dense FlashAttention medians by HetA medians under the A100 protocol of Table~\ref{tab:heta_latency}: $121\times704\times1280$, BF16, 80\% local queries, and 15 warmed repeats including routing and execution. The $11^3$ timings are reused. FLOPs retain the original arithmetic count. The default $11^3$ window has the highest semantic score.}
\label{tab:kernel_size_ablation}

\end{minipage}
\end{table*}

\section{Results}
\label{sec:main_results}

\begin{table*}[t]
\centering
\small
\setlength{\tabcolsep}{5pt}
\caption{\textbf{VBench2.0 comparison by category.} Reproduced Wan2.2-5B variants use $121\times704\times1280$; reported model results retain their native evaluation settings. Bold/underline rank the three-step and two-step DMD variants.}
\label{tab:vbench2_sota}
\begin{tabular}{lcccccc}
\toprule
Model & Total$\uparrow$ & \makecell{Human\\Fidelity$\uparrow$} & Creativity$\uparrow$ & Controllability$\uparrow$ & \makecell{Common\\Sense$\uparrow$} & Physics$\uparrow$ \\
\midrule
Wan2.1 1.3B & 56.0 & 80.7 & 48.7 & 34.0 & 63.4 & 53.8 \\
Wan2.2 5B & 58.1 & 79.4 & 57.7 & 36.8 & 64.1 & 51.3 \\
HunyuanVideo & 55.3 & 82.4 & 41.8 & 28.6 & 63.4 & 60.2 \\
CogVideoX-1.5 & 53.4 & 72.1 & 43.7 & 29.6 & 63.2 & 48.2  \\
\midrule
Dense DMD2 (Wan2.2-5B) & 57.95 & 76.9 & 54.2 & \textbf{37.9} & 59.9 & 60.9 \\
Dense DMD3 (Wan2.2-5B) & \textbf{60.80} & \textbf{80.7} & \textbf{58.7} & \underline{36.2} & \textbf{64.5} & \textbf{63.9}  \\

\rowcolor{methodblue}
HetA-DiT (Wan2.2-5B) & \underline{58.92} & \underline{79.8} & \underline{55.0} & 35.9 & \underline{62.5} & \underline{61.4}  \\
\bottomrule
\end{tabular}
\end{table*}

\begin{table*}[!t]
\centering
\begin{minipage}[t]{0.49\textwidth}
\vspace{0pt}
\centering\small
\setlength{\tabcolsep}{5pt}
\begin{tabularx}{\linewidth}{@{}lCCC@{}}
\toprule
Global allocation & Total$\uparrow$ & Quality$\uparrow$ & Semantic$\uparrow$ \\
\midrule
All-random & 82.78 & 83.37 & 80.45 \\
HetA, no swaps & 83.04 & 83.48 & 81.25 \\
HetA, 3\% swaps & \underline{83.18} & \underline{83.63} & \underline{81.39} \\
\rowcolor{methodblue}
HetA, 7\% swaps & \textbf{83.28} & \textbf{83.71} & \textbf{81.52} \\
\bottomrule
\end{tabularx}
\caption{\textbf{Random-swap ablation on Wan2.2-5B.} Full VBench with an $11^3$ neighborhood. Swaps exchange equal local/global query counts at each block. All-random redraws the entire global allocation.}
\label{tab:main_swap_ablation}
\end{minipage}\hfill
\begin{minipage}[t]{0.49\textwidth}
\vspace{0pt}
\centering\small
\setlength{\tabcolsep}{4pt}
\begin{tabularx}{\linewidth}{@{}lCC@{}}
\toprule
Property & Adapted AHA & HetA \\
\midrule
Signal & Current-layer gates & Denoiser uncertainty \\
Supervision & Task + sparsity & Uncertainty DMD \\
Head allocation & Head-specific & Shared token route \\
Block routing & Recomputed & Previous-step reuse \\
\bottomrule
\end{tabularx}
\caption{\textbf{AHA and HetA routing designs.} Both use $11^3$ context, 7\% swaps and matched DMD adaptation. The comparison evaluates the complete routing.}
\label{tab:main_aha_design}
\end{minipage}
\end{table*}

\subsection{Generation Quality and Human Preference}
HetA scores \textbf{83.28 versus 82.94} for dense DMD3 on full VBench, despite approximately 80\% of its queries being local (Table~\ref{tab:vbench_sota}). It also exceeds the adapted AHA baseline on both quality and semantics. On Wan2.1-1.3B (Table~\ref{tab:vbench_sota_wan2.1}), HetA (83.10) matches the reproduced Wan model and exceeds VSA. Training-free caching performs less well in this few-step setting, consistent with its reliance on repeated-step redundancy~\citep{zou2026disca}. The Wan2.1-14B comparison below extends the quality evaluation to a larger dense-attention student.

On VBench2.0 (Table~\ref{tab:vbench2_sota}), HetA exceeds dense DMD2 in total and in every category except Controllability, scoring 58.92 versus 57.95, and remains close to dense DMD3 (60.80). It improves Instance Preservation, Motion Order Understanding and Complex Landscape over dense DMD3, consistent with strong subject preservation.

In the blinded human study (Table~\ref{tab:human_preference}), HetA wins 50\% of comparisons against DMD2 versus 24\% for dense, and 68\% against Attention Surgery versus 19\%. Against dense DMD3, preferences are closely balanced (32\% HetA, 32\% ties, 36\% dense). Figure~\ref{fig:qual} shows prompt-matched qualitative examples.

\subsection{Scaling to Wan2.1-14B}
\label{sec:wan14}
HetA reaches \textbf{84.01 total} on full VBench, compared with 84.63 for dense three-step DMD3 (Table~\ref{tab:wan14_quality}). Its semantic score improves from 81.97 to 82.29, while quality is 84.44 versus 85.29. This extends uncertainty-guided attention to a 14B student using the same approximately 80\% local-query policy as 5B.

\begin{table}[t]
\centering
\begin{minipage}[t]{0.49\linewidth}
\vspace{0pt}
\centering\footnotesize
\setlength{\tabcolsep}{3pt}
\begin{tabularx}{\linewidth}{@{}lCCC@{}}
\toprule
Wan2.1-14B & Total$\uparrow$ & Quality$\uparrow$ & Semantic$\uparrow$ \\
\midrule
Dense DMD3 & \textbf{84.63} & \textbf{85.29} & 81.97 \\
\rowcolor{methodblue}HetA-DiT & 84.01 & 84.44 & \textbf{82.29} \\
\bottomrule
\end{tabularx}
\caption{\textbf{Full VBench on Wan2.1-14B.} Both methods use three steps at $81\times480\times832$ with the original decoder. HetA uses 500 uncertainty-stage and 3,000 adaptation-stage generator updates; its training and inference noise shift is 5.}
\label{tab:wan14_quality}

\end{minipage}\hfill
\begin{minipage}[t]{0.49\linewidth}
\vspace{0pt}
\centering\footnotesize
\setlength{\tabcolsep}{3pt}
\begin{tabularx}{\linewidth}{@{}lCC@{}}
\toprule
Cohort & \makecell{Spearman\\(95\% CI)} & \makecell{Capture/random\\(95\% CI)} \\
\midrule
Initial eight & \makecell{0.319\\{}[0.227, 0.415]} & \makecell{$1.26\times$\\{}[1.20, 1.33]} \\
Disjoint eight & \makecell{0.359\\{}[0.289, 0.439]} & \makecell{$1.29\times$\\{}[1.21, 1.38]} \\
\bottomrule
\end{tabularx}
\caption{\textbf{Uncertainty tracks global-context sensitivity (Wan2.2-5B).} Previous-step uncertainty versus dense/local output change; 95\% intervals resample prompts.}
\label{tab:uncertainty_summary}

\end{minipage}
\end{table}

We begin with a LoRA-distilled Wan2.1-14B student and train its uncertainty heads for 500 generator updates with dense attention, approximately 7\% of the 5B uncertainty schedule. We then freeze uncertainty and adapt only the existing attention LoRA parameters for 3,000 generator updates, with one fake-score update per generator update. Training uses noise shift 5, matching the dense 14B recipe and the evaluation scheduler. Full VBench uses 944 GPT-enhanced prompts and five samples per prompt, yielding 4,720 scored videos. Generation uses three steps, classifier-free guidance 1, 81 frames at $480\times832$, and the original Wan decoder.

The 0.62-point total gap is concentrated in the quality aggregate, while the semantic aggregate exceeds dense DMD3. The main comparison therefore reports both aggregates. Appendix Figure~\ref{fig:wan14_qualitative} shows prompt-matched examples from this evaluated checkpoint; full optimization and inference details are in Appendix~\ref{sec:wan14_training}.

\subsection{Routing Analysis and Ablations}
\label{sec:ablations}
With matched initialization, neighborhood, swaps and adaptation budget, uncertainty routing exceeds uniform routing by \textbf{0.60 VBench points} (95\% paired prompt interval $[+0.20,+0.97]$). It exceeds adapted AHA by 0.72 points ($[+0.36,+1.08]$), with slightly fewer threshold-selected global queries.

\begin{table}[t]
\centering
\begin{minipage}[t]{0.49\linewidth}
\vspace{0pt}
\centering
\includegraphics[width=\linewidth]{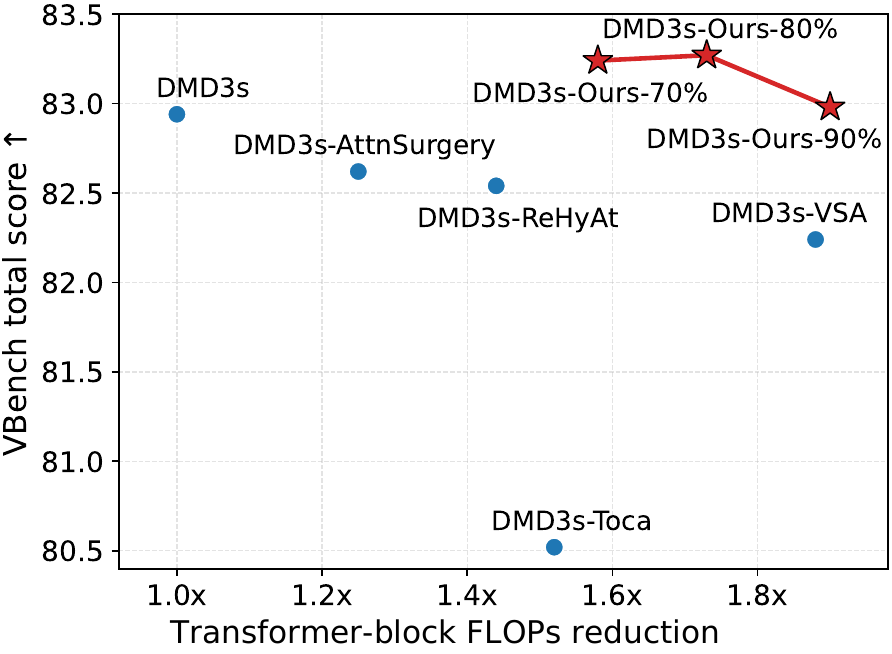}
\captionof{figure}{\textbf{Quality-computation trade-off on Wan2.2-5B.} The horizontal axis is DeepSpeed-profiled Transformer-block FLOPs reduction. HetA's 70\%, 80\% and 90\% local-query settings retain competitive VBench scores at lower arithmetic cost.}
\label{fig:quality_eff_tradeoff}
\end{minipage}\hfill
\begin{minipage}[t]{0.49\linewidth}
\vspace{0pt}
\centering\footnotesize
\setlength{\tabcolsep}{3pt}
\renewcommand{\arraystretch}{1.3}
\begin{tabular}{@{}lccrrr@{}}
\toprule
Routing & Window & Swaps & Total$\uparrow$ & Qual.$\uparrow$ & Sem.$\uparrow$ \\
\midrule
Local-only & $11^3$ & - & 82.85 & \underline{83.56} & 79.99 \\
Random & $11^3$ & All$^*$ & 82.78 & 83.37 & 80.45 \\
Uniform + swaps & $11^3$ & 7\% & 82.68 & 83.42 & 79.76 \\
\midrule
HetA, no swaps & $11^3$ & 0\% & \underline{83.04} & 83.48 & \underline{81.25} \\
\rowcolor{methodblue}HetA + swaps$^\dagger$ & $11^3$ & 7\% & \textbf{83.28} & \textbf{83.71} & \textbf{81.52} \\
\midrule
\multicolumn{6}{l}{Different-window references} \\ \midrule
Fixed uniform & $13^3$ & 0\% & 83.13 & 83.76 & 80.58 \\
HetA + swaps & $13^3$ & 7\% & 83.36 & 83.99 & 80.82 \\
\bottomrule
\end{tabular}
\caption{\textbf{Routing ablation on full VBench.} The 7\%-swap $11^3$ rows match neighborhood and adaptation budget. Bold/underline rank $11^3$ rows. $^*$Global assignment is redrawn randomly each block. $^\dagger$Paired prompt interval for HetA vs.\ uniform: $[+0.20,+0.97]$. Compared with local-only, default HetA improves semantics by 1.53 points.}
\label{tab:routing_ablation}

\end{minipage}
\end{table}

Local-only routing scores 82.85 total but 79.99 semantic (Table~\ref{tab:routing_ablation}). Uncertainty routing raises the semantic score by 1.53 points over this reference, versus at most 0.59 for the non-uncertainty alternatives, so the global budget pays off most when targeted by uncertainty. The fixed-uniform reference uses a $13^3$ neighborhood without swaps; at that neighborhood, HetA scores 83.36 versus 83.13.

\paragraph{Routing design and execution.}
Our AHA adaptation retains its current-layer token/head gates, straight-through estimator and sparsity penalty, replacing the causal language-model window with a noncausal $11^3$ neighborhood and its task loss with DMD. Beyond the routing signal (Table~\ref{tab:main_aha_design}), HetA shares one token list across heads and reuses preceding-step predictions across blocks, avoiding repeated gate evaluation and packing. With the same local kernel and optimized global packing (supplement), AHA's grouped heads require six global calls per sparse 5B block versus one for HetA. At exactly equal global-query counts, two paired prompt/seed cases take 8.13\,s of DiT time for HetA versus 9.98\,s for optimized AHA ($1.22\times$), including the dense first step and live routing.

\paragraph{Sparsity operating points.}
Figure~\ref{fig:quality_eff_tradeoff} compares the quality-computation trade-off with efficient-attention baselines. The 70\%, 80\% and 90\% local-query settings score 83.24, 83.28 and 82.99 on VBench, with swap rates 10\%, 7\% and 5\%.

\paragraph{Why random swaps help.}
Increasing swaps from 0\% to 3\% to 7\% improves total and semantic scores (Table~\ref{tab:main_swap_ablation}). Equal local/global exchanges give initially local tokens global context while preserving the attention budget. Under independent uniform draws, a base-local query has probability $1-(1-0.07/0.8)^{30}\approx0.94$ of promotion across 30 blocks. Fully random allocation scores lower, supporting uncertainty-guided allocation with broader coverage.

\paragraph{Local window size.}
Table~\ref{tab:kernel_size_ablation} varies the 3D window at 7\% swaps. Expanding from $9^3$ to $13^3$ raises total VBench from 83.22 to 83.36, while attention speedup over dense decreases from $2.48\times$ to $1.80\times$ and block speedup from $1.37\times$ to $1.24\times$. The default $11^3$ window has the highest semantic score (81.52).

\subsection{Does Uncertainty Predict Global-Context Sensitivity?}
\label{sec:uncertainty_results}
Running the trained 5B denoiser densely and locally on identical noisy inputs, previous-step uncertainty correlates with the whole-denoiser output change when global context is removed: mean Spearman correlation is \textbf{0.319 and 0.359} in the initial and disjoint prompt cohorts (Table~\ref{tab:uncertainty_summary}). This directly tests the signal available when HetA selects a route. On disjoint prompts, the selected 18.25\% of eligible tokens capture 23.60\% of squared output discrepancy, $1.29\times$ random expectation ($1.26\times$ on the initial cohort). Each token's discrepancy is the RMS difference of its predicted update over its output patch; each cohort has eight prompts, two seeds and both later steps, with intervals over prompt clusters. The appendix details the protocol and visualizes the sensitivity maps.

\subsection{Attention, DiT and Generation Latency}
\label{sec:latency_results}
\begin{table*}[t]
\centering
\small
\setlength{\tabcolsep}{5pt}
\begin{tabular}{@{}lcrrrr@{}}
\toprule
\multirow{2}{*}{Model} & \multirow{2}{*}{Video size ($T\times H\times W$)} & \multicolumn{2}{c}{Latency speedup} & \multicolumn{2}{c}{FLOPs reduction} \\
\cmidrule(lr){3-4}\cmidrule(lr){5-6}
& & Attention & Block & Attention & Block \\
\midrule
Wan2.1-1.3B & $81\times480\times832$ & $1.75\times$ & $1.32\times$ & $4.12\times$ & $2.12\times$ \\
Wan2.2-5B & $121\times704\times1280$ & $2.05\times$ & $1.30\times$ & $4.37\times$ & $1.69\times$ \\
Wan2.2-14B expert & $121\times480\times832$ & $2.77\times$ & $1.57\times$ & $4.60\times$ & $1.94\times$ \\
Wan2.2-14B expert & $121\times704\times1280$ & $3.78\times$ & $2.23\times$ & $4.81\times$ & $2.65\times$ \\
\bottomrule
\end{tabular}
\caption{\textbf{Attention and Transformer-block efficiency.} Latency compares the Triton implementation with dense FlashAttention using synthetic inputs, 80\% local queries, fresh 7\% swaps, and medians of 15 warmed A100 repeats including routing, gathering and scattering. FLOPs count dominant matrix products under the same query budget; non-matmul operations and tile padding are excluded from arithmetic counts but included in latency. The 14B rows use one pretrained expert.}
\label{tab:heta_latency}
\end{table*}

\begin{table*}[!t]
\centering
\footnotesize
\setlength{\tabcolsep}{3pt}
\begin{tabular*}{\textwidth}{@{\extracolsep{\fill}}llrrrrrr@{}}
\toprule
\multirow{2}{*}{Model} & \multirow{2}{*}{Video size ($T\times H\times W$)} & \multicolumn{3}{c}{3-step DMD} & \multicolumn{3}{c}{50 steps + CFG (100 NFEs)} \\
\cmidrule(lr){3-5}\cmidrule(lr){6-8}
& & Tiled VAE & Untiled VAE & LightTAE & Tiled VAE & Untiled VAE & LightTAE \\
\midrule
Wan2.1-1.3B & $81\times480\times832$ & $1.069\times$ & $1.103\times$ & $1.183\times$ & $1.291\times$ & $1.300\times$ & $1.308\times$ \\
Wan2.2-5B & $121\times704\times1280$ & $1.034\times$ & $1.061\times$ & $1.163\times$ & $1.237\times$ & $1.252\times$ & $1.266\times$ \\
Wan2.2-14B expert & $121\times480\times832$ & $1.246\times$ & $1.279\times$ & $1.319\times$ & $1.551\times$ & $1.554\times$ & $1.556\times$ \\
Wan2.2-14B expert & $121\times704\times1280$ & $1.455\times$ & $1.491\times$ & $1.537\times$ & $2.055\times$ & $2.058\times$ & $2.062\times$ \\
\bottomrule
\end{tabular*}
\caption{\textbf{End-to-End Generation latency improvement across decoder costs and schedules.} HetA speedup for DiT plus decoding on A100. Entries compose separately measured components, except the directly measured 5B tiled-VAE pipeline; all 100-NFE entries project runtime with two dense first-step CFG evaluations. The 1.3B and 14B/480p DiT costs stack measured block times; the other rows use complete DiT timings. The 14B rows benchmark one pretrained expert with controlled routes. Tiled VAE uses the original small-tile policy; both methods use the same decoder in each column. Detailed component costs and scope are in the appendix.}
\label{tab:e2e_latency}
\end{table*}

Separate local/global execution produces up to \textbf{$3.78\times$ attention and $2.23\times$ block speedup} over dense FlashAttention in the 14B/704p operator benchmark (Table~\ref{tab:heta_latency}), and $2.05\times$ and $1.30\times$ at 5B/704p. The 14B rows benchmark a pretrained expert with controlled routes.

On trained Wan2.2-5B, median three-step DiT latency across 32 prompts and two seeds is 8.15\,s for HetA versus 9.52\,s for DMD3, including the dense first step, uncertainty prediction and all routing costs; routing and swaps take only 0.097\,s. At this short schedule, both methods share approximately 31.4\,s of original VAE decoding, which dominates generation time and can be reduced substantially.

\paragraph{End-to-end latency across operating points.}
Table~\ref{tab:e2e_latency} shows how decoder cost and denoising schedule affect generation speedup. Gains grow with model size, resolution and the number of denoising steps. Faster decoding also increases them: disabling small spatial tiles in the original VAE reduces 5B decoding from 30.88\,s to 14.02\,s, and the LightTAE decoder~\citep{lightx2v2026autoencoders} reconstructs identical HetA latents at 27.17\,dB PSNR, SSIM 0.844~\citep{wang2004ssim} and LPIPS 0.061~\citep{zhang2018lpips} relative to the original decoder; full VBench and human-preference scores use the original decoder. The 50-step CFG columns project 100 forward evaluations, including two dense first-step evaluations, reaching up to $2.06\times$ for the 14B expert. The 1.3B and 14B/480p rows stack block medians; the appendix specifies their scope.

\section{Conclusion}
HetA-DiT allocates global video attention using a supervised uncertainty signal while updating all tokens. Sharing routes and reusing preceding-step predictions makes this design compatible with few-step DMD. On 5B, HetA is competitive with dense three-step DMD on VBench with approximately 80\% of queries local, and it extends to 1.3B and 14B. Matched routing comparisons and sensitivity diagnostics support uncertainty guidance, and separate Triton execution converts query sparsity into latency savings. Speedups grow with model size and sampling workload, as attention dominates cost. Learned uncertainty could also steer other adaptive budgets, such as step counts or caching.

\clearpage
\bibliography{references}
\clearpage
\appendix

\section{Training Details}
\subsection{Training of HetA-DiT.}
We provide additional details on the two training stages of HetA-DiT. The model is trained using the standard DMD pipeline, with the objective functions for each stage described in the main paper. The first training stage consists of 15k fine-tuning iterations using a per-GPU batch size of 2 across 4 GPUs. The student learning rate is set to $1 \times 10^{-5}$, while the critic learning rate is set to $5 \times 10^{-6}$. We employ a constant learning rate schedule with a 10-step warm-up phase. Optimization is performed using AdamW with a weight decay of 0.01, $\beta_1 = 0.9$, and $\beta_2 = 0.999$. Training is conducted in mixed precision using bfloat16. For timestep sampling, we draw timesteps from a shifted uniform distribution with a shift parameter of $s=5$. The teacher classifier-free guidance scale is 5; few-step student inference uses guidance scale 1. These iteration counts include alternating generator and fake-score updates: 7,500 generator updates in Stage 1. The second training stage consists of 6k alternating fine-tuning iterations (3,000 generator updates) using a per-GPU batch size of 1 across 4 GPUs. Timesteps are again sampled from a shifted uniform distribution; however, the shift parameter is set to $s=1$. All other hyperparameters remain unchanged.

\subsection{Training of baselines.}
For Attention Surgery, we replaced 20 transformer blocks with hybrid blocks with the hybrid rate of 8 (20$\times$R8 configuration).
Following the recipe of \citet{ghafoorian2026attentionsurgery}, the training comprised three stages: block-wise distillation, flow-matching finetuning, and DMD2 distillation.
The same 3-stage pipeline was applied to ReHyAt but in that case all 30 blocks of Wan2.2 video transformer have been replaced.
Every chunk consisted of 3 latent frames, and chunk overlap was set to a single frame.
Both Attention Surgery and ReHyAt used polynomials of degree 2 in their linear attention branch.

In our VSA training, we mostly followed the recipe provided by FastVideo library and set the sparsity rate to 0.8 while spatiotemporal tile size was $4 \times 4 \times 4$.
In contrast, Jenga is a non-trainable block-sparse attention method.
We applied it on top of our baseline DMD checkpoint with sparsity ratio of 0.8, threshold probability of 0.9, and tile size equal to 64. 

Similar to Jenga, ToCa is a training-free caching method. To make ToCa more suited for the step-distilled regime, we run it only on \(1\) out of \(3\) steps: the first and last denoising steps are evaluated fully, while ToCa caching is applied only on the middle step. We set the token refresh ratio to \(0.6\), meaning that on the cached step \(60\%\) of FFN tokens are recomputed and the remaining tokens reuse cached activations; with the ToCa text-to-video module scheduler, this corresponds to an effective FFN refresh ratio of approximately \(89.7\%\) and an FFN cache ratio of approximately \(10.3\%\).

\subsection{Wan2.1-14B adaptation}
\label{sec:wan14_training}
The main-paper 14B result uses a three-step LoRA-distilled Wan2.1-14B student. Stage 1 updates only uncertainty heads for 500 generator updates using dense attention; the student and fake-score weights are frozen. Stage 2 freezes uncertainty and updates the existing attention LoRA parameters with rank and scaling factor 512. The evaluated checkpoint contains 3,000 student and 3,000 fake-score updates.

Adaptation uses four H100 GPUs, BF16, one sample per GPU and no gradient accumulation. AdamW uses student/fake-score learning rates $10^{-5}/5\times10^{-6}$, $\beta_1=0.9$, $\beta_2=0.999$, weight decay 0.01, ten-update warmup followed by constant learning rates, and gradient clipping at norm 1. Teacher guidance is 3. Training noise, re-noising and fake-score noise use shift 5. Prompts are reshuffled each epoch, and resumed training continues at the next unconsumed minibatch. Routing retains threshold $c=0.9$, an $11^3$ neighborhood, 7\% per-block swaps and a dense first denoising step.

Evaluation uses 944 GPT-enhanced prompts, all 16 VBench dimensions and five samples per prompt. HetA inference seeds are 2027--2031. Videos contain 81 frames at $480\times832$ and 16 fps; inference uses three Euler flow-matching steps, noise shift 5, classifier-free guidance 1, BF16 and the original Wan decoder. We reuse the existing dense DMD3 evaluation with matching geometry, steps, guidance, shift and prompts; inference seeds are not paired. Table~\ref{tab:wan14_quality} reports the results in the main paper.

\section{Routing Controls and AHA Adaptation}
\label{sec:routing_controls}
\subsection{Matched uniform routing}
The uniform-with-swaps control starts from the same dense Wan2.2-5B student, uses the same $11\times11\times11$ neighborhood and 7\% random swaps as HetA, and receives 3,000 generator updates with one fake-score update per generator update. Its full-VBench score is 82.68 total, 83.42 quality and 79.76 semantic, compared with HetA's 83.28, 83.71 and 81.52. The fixed-uniform reference instead uses a $13^3$ window and no swaps; the matched comparison uses the 7\%-swap control.

\subsection{Paired prompt intervals}
Full VBench uses 944 canonical GPT-enhanced prompts and five samples per prompt, or 4,720 scored videos per method. We draw 20,000 bootstrap samples of the 944 prompts with replacement, sharing each draw across methods. Each selected prompt carries its five samples and all of its evaluation-dimension memberships. Each draw recomputes the official per-dimension averages, normalization and weighted quality and semantic aggregates, followed by their $4{:}1$ weighted total.

The 95\% paired interval for HetA minus uniform-with-swaps is $[+0.20,+0.97]$ total-score points. For HetA minus AHA, it is $[+0.36,+1.08]$. Both intervals exclude zero.

\subsection{AHA adaptation}
\label{sec:aha_adaptation}
We implement All-or-Here Attention~\citep{luo2025aha} on Wan2.2-5B, retaining its current-layer linear/sigmoid token/head gates, selected-path straight-through estimator, task loss and sparsity penalty. For video diffusion, its causal 1D language-model window is replaced with a noncausal $11^3$ neighborhood and its task loss with DMD. Both AHA and HetA start from the dense student and retain a dense first denoising step. AHA keeps its top-scoring 20\% of eligible token/head positions, followed by 7\% swaps. HetA's threshold selects approximately 18\% before swaps. Both receive the full matched adaptation budget.

\subsection{Optimized AHA execution}
We separately optimize AHA's head-specific execution with the same local Triton kernel and FlashAttention for global queries. We test padding all heads to the largest global-query allocation and sorting heads into groups of four to reduce padding. Both preserve the selected query sets and pass independent attention-reference checks. For native trained 5B denoising routes, grouping reduces median global-query padding from $2.51\times$ to $1.17\times$ selected slots, using six global FlashAttention calls per sparse block. HetA uses one common query list across heads and one global call, while reusing uncertainty across blocks.

With exactly equal global-query counts, complete three-step 5B denoising takes 8.13\,s for HetA and 9.98\,s for optimized AHA ($1.22\times$ paired speedup). This uses two paired prompt/seed cases on the same A100 in BF16 at batch one. It includes the dense first step, live gate/uncertainty prediction, swaps, gathering/scattering and scheduler updates; text encoding, decoding, loading and warmup are excluded.

\begin{table}[t]
\centering\small
\setlength{\tabcolsep}{3pt}
\begin{tabular}{lcc}
\toprule
Model / resolution & HetA & Best tested AHA variant \\
\midrule
1.3B / $480\times832$ & $1.61/1.27$ & $1.04/1.03$ \\
5B / $704\times1280$ & $1.90/1.28$ & $1.00/1.00$ \\
14B expert / $480\times832$ & $2.72/1.57$ & $1.86/1.36$ \\
14B expert / $704\times1280$ & $3.75/2.22$ & $3.27/2.08$ \\
\bottomrule
\end{tabular}
\caption{Attention / complete-block speedup over dense FlashAttention with equal 20\% global-query budgets, 7\% swaps and nine interleaved A100 repetitions. Only 5B uses trained AHA gates; other rows use seeded routing controls. This nine-repeat interleaved HetA/AHA protocol is separate from the 15-repeat Triton protocol of the main-paper operator table, so absolute speedups differ slightly (e.g., 5B attention $1.90\times$ here versus $2.05\times$ there).}
\label{tab:aha_execution}
\end{table}

The approximately $1.00\times$ AHA result at 5B includes gate evaluation, per-head routing and packing/dispatch. Reduced arithmetic does not necessarily produce a runtime gain when these costs dominate.

\section{Triton Implementation and Latency Protocols}
\label{sec:implementation_timing}
\subsection{Separate execution and route management}
After fresh swaps at each block, selected local queries are evaluated by a dedicated Triton kernel over clipped 3D neighborhoods. Global queries are gathered into one shared list across heads, attend to all original keys and values through FlashAttention, and are scattered back alongside local outputs. Original coordinates are preserved. The 5B local kernel uses 16 query rows in $2\times2\times4$ spatial/temporal tiles and chunks of 64 keys. Neighborhood geometry is cached; query membership is rebuilt after swaps. All rebuilding, packing, padding and scattering costs are included in latency.

We inspect saved routes from one example uncertainty-guided generation. Across its 60 sparse blocks, 74.35\% of queries are local; because the threshold is adaptive, this fraction varies with content around the nominal 80\% used in our benchmarks. In a hypothetical mixed flat 64-query layout, 99.9961\% of tiles would contain a global query and therefore load all keys; mixed spatial 16-query tiles reach 91.88\%. The separate local/global calls avoid this coupling.

\subsection{Operator benchmark}
The attention/block measurements use A100-SXM4-80GB, PyTorch 2.8.0+cu128, Triton 3.4.0, BF16, batch one, synthetic activations, 80\% local queries and fresh 7\% swaps. We test sixteen kernel configurations per workload, check sampled query outputs against independent FP32 attention, and report synchronized medians over 15 warmed repeats. The dense baseline forces FlashAttention with all queries attending to all original keys and values. A separately tuned dense Triton implementation is slower on the four tested workloads.

Attention timing operates on projected queries, keys and values. Complete-block timing additionally includes self-attention projections, normalization and rotary embeddings, cross-attention, FFN, and fresh route swaps. The 1.3B control uses a $11\times15\times15$ local window and 81 frames; 5B and 14B use $11^3$ and 121 frames. The 14B rows use one pretrained Wan2.2 expert with controlled routes.

The main window-size ablation additionally times $9^3$ and $13^3$ neighborhoods on Wan2.2-5B with the same video geometry, checkpoint block, synthetic states, routing budget and timing procedure. Each window tests the same sixteen execution configurations against sampled FP32 attention and uses its selected configuration for 15 warmed repeats. Its speedup divides the dense FlashAttention median by the corresponding HetA median from the same benchmark workload. The $11^3$ row reuses the original operator measurements. Quality scores come from the models trained with each window.

\begin{table*}[t]
\centering\small
\setlength{\tabcolsep}{5pt}
\begin{tabular}{lrrrrrrrr}
\toprule
Model / resolution & \multicolumn{2}{c}{Attention (ms)} & \multicolumn{2}{c}{Block (ms)} & \multicolumn{2}{c}{Matmul FLOPs reduction} & Query tile & KV chunk \\
\cmidrule(lr){2-3}\cmidrule(lr){4-5}\cmidrule(lr){6-7}
& Dense & HetA & Dense & HetA & Attention & Block & & \\
\midrule
1.3B / $480\times832$ & 34.16 & 19.52 & 59.17 & 44.95 & $4.12\times$ & $2.12\times$ & $2\times4\times4$ & 128 \\
5B / $704\times1280$ & 46.86 & 22.89 & 101.18 & 77.86 & $4.37\times$ & $1.69\times$ & $2\times2\times4$ & 64 \\
14B / $480\times832$ & 241.85 & 87.22 & 419.20 & 266.26 & $4.60\times$ & $1.94\times$ & $2\times2\times4$ & 64 \\
14B / $704\times1280$ & 1233.98 & 326.39 & 1635.52 & 733.87 & $4.81\times$ & $2.65\times$ & $2\times4\times4$ & 128 \\
\bottomrule
\end{tabular}
\caption{Absolute medians behind the main Triton table, with dominant-matmul arithmetic reductions and selected tile sizes. FLOPs exclude non-matmul operations and masked tile padding; latency includes routing and execution costs.}
\label{tab:operator_details}
\end{table*}

For $N$ tokens, hidden width $C$ and $P$ allowed query/key pairs per head, the two attention matrix products cost $4CP$ FLOPs when multiply-add counts as two operations. The block count adds self-attention projections, cross-attention projections/products and both FFN projections. HetA counts exact clipped neighborhoods for the representative route; dense uses $P=N^2$. These arithmetic counts exclude softmax, norms, activations, rotary embeddings, routing and masked tile padding. Consequently, the fixed-budget 5B block reduction is $1.69\times$; the $1.73\times$ in the main trade-off figure is a DeepSpeed profile at the trained threshold setting, which also counts non-matmul operations.

\subsection{Complete trained-model timing}
The complete 5B protocol uses 32 GPT-enhanced prompts and two seeds each at $121\times704\times1280$, interleaving methods on the same GPU. DiT time is a CUDA stream interval from the first Transformer invocation through the last scheduler update, including uncertainty, selection, swaps and dispatch gaps. Pipeline timing is synchronized wall time through decoded frames, including text encoding and VAE/postprocessing. Loading, compilation, warmup and MP4 encoding are excluded. Prompt-bootstrap intervals retain both seeds together.

\begin{table*}[t]
\centering\small
\setlength{\tabcolsep}{10pt}
\begin{tabular}{lrrrr}
\toprule
Method & Complete DiT (s) & Full pipeline (s) & VBench & VBench2.0 \\
\midrule
Dense DMD2 & \textbf{6.34} & \textbf{37.83} & 82.48 & 57.95 \\
Dense DMD3 & 9.52 & 41.01 & 82.94 & \textbf{60.80} \\
\rowcolor{methodblue}HetA3 & \underline{8.15} & \underline{39.65} & \textbf{83.28} & \underline{58.92} \\
\bottomrule
\end{tabular}
\caption{\textbf{Measured 5B generation latency and quality.} Medians over 32 prompts and two seeds on A100 at $121\times704\times1280$. Pipeline includes text encoding, DiT, original VAE and frame postprocessing. Compilation, loading and MP4 encoding are excluded. Quality uses the same attention rule, validated against a reference.}
\label{tab:complete_generation}
\end{table*}

\begin{table}[t]
\centering\small
\setlength{\tabcolsep}{4pt}
\begin{tabular}{lrr}
\toprule
Component / interval (s) & Dense DMD3 & HetA3 \\
\midrule
Complete DiT sequence & 9.52 & 8.15 \\
Self-attention, incl. gather/scatter & 5.601 & 4.203 \\
First dense denoising step & $\approx3.18$ & $\approx3.18$ \\
Routing / swaps / rebuilding & - & 0.097 \\
\bottomrule
\end{tabular}
\caption{Nested timing intervals for trained Wan2.2-5B. Component values are already included in their parent intervals; the rows must not be added together.}
\label{tab:timing_breakdown}
\end{table}

\subsection{Composed generation latency}
The main end-to-end table reuses saved A100 DiT evidence and separately measured decoders. Let $D$ and $H$ be dense and heterogeneous forward costs and $V$ the shared decoder cost. Three-step composition is
\begin{equation}
t_d=3D+V,\qquad t_h=D+2H+V.
\end{equation}
For 50 sampling steps with classifier-free guidance (100 NFEs), both initial CFG evaluations remain dense:
\begin{equation}
t_d=100D+V,\qquad t_h=2D+98H+V.
\end{equation}
These totals model DiT plus decoding.

\begin{table*}[t]
\centering\small
\setlength{\tabcolsep}{5pt}
\begin{tabular}{lrrrrrr}
\toprule
Model / video size & Layers & Dense $D$ & HetA $H$ & Tiled VAE & Untiled VAE & LightTAE \\
\midrule
Wan2.1-1.3B / $81 \times 480 \times 832$ & 30 & 1.775 & 1.348 & 7.911 & 3.825 & 0.181 \\
Wan2.2-5B / $121 \times 704 \times 1280$ & 30 & 3.172 & 2.491 & 30.880 & 14.015 & 0.204 \\
Wan2.2-14B expert / $121 \times 480 \times 832$ & 40 & 16.768 & 10.650 & 11.730 & 5.729 & 0.267 \\
Wan2.2-14B expert / $121 \times 704 \times 1280$ & 40 & 66.626 & 31.605 & 24.128 & 12.867 & 0.581 \\
\bottomrule
\end{tabular}
\caption{Saved component costs in seconds used for the main generation-speedup table. $D,H$ are effective single-forward costs; decoder columns are independently measured warmed medians. The 1.3B and 14B/480p DiT values stack blocks; 5B uses complete three-step medians and 14B/704p complete expert forwards.}
\label{tab:composition_components}
\end{table*}

The 5B three-step row uses complete 64-case denoising medians. Its effective projection costs are $D=t_{d,3}/3$ and $H=(t_{h,3}-D)/2$. The 14B/704p row uses six paired complete 40-block single-expert forwards, including embeddings and output projection, with controlled 80\% local routes. The 1.3B and 14B/480p rows multiply saved block medians by verified layer counts (30 and 40). These two estimates exclude input/output embedding and uncertainty-head overhead, as marked in the main table. The 14B estimates use a single expert.

Original-decoder tiling uses 256-pixel tiles with stride 192; untiled decoding keeps the same weights and disables spatial tiling. Decoder measurements use BF16, batch one, fixed seeds and repeated warmed calls. Wan2.2-14B uses the Wan2.1 VAE, whereas 5B uses its own decoder.

LightTAE~\citep{lightx2v2026autoencoders} uses matching Wan2.1/Wan2.2 decoder weights and default sequential execution. Nine warmed A100 calls are measured for each VAE/geometry pair. We do not extrapolate decoder cost across model formats or resolutions.

\subsection{LightTAE reconstruction fidelity}
Twenty 5B prompts and one seed per prompt were fixed before generation. We cache final three-step HetA latents at $121\times704\times1280$ with student guidance scale 1 and decode those identical latents with the original untiled VAE and LightTAE. Metrics use every native-resolution frame before MP4 compression. PSNR is computed from whole-video mean squared error in FP64 and then averaged over prompts; SSIM~\citep{wang2004ssim} and AlexNet LPIPS~\citep{zhang2018lpips} are averaged over frames and then prompts. Mean values are 27.17\,dB, 0.844 and 0.0609, respectively. Full VBench and human-preference evaluations use the original VAE.

\section{Additional Quality Results and Ablations}
\begin{table*}[t]
\centering
\small
\caption{All VBench dimensions for dense DMD3 and the submitted HetA sparsity settings. Values are percentages before normalization; aggregate scores use the official weights. Local fractions are approximate; swap fractions are 10\%, 7\% and 5\%, respectively.}
\label{tab:vbench_dimensions}
\begin{tabular}{lrrrr}
\toprule
Dimension$\uparrow$ & Dense DMD3 & HetA 70\% & HetA 80\% & HetA 90\% \\
\midrule
Subject consistency & 95.10 & 95.14 & 94.89 & 95.05 \\
Background consistency & 95.31 & 95.97 & 95.77 & 95.83 \\
Temporal flickering & 98.86 & 98.76 & 98.77 & 98.79 \\
Motion smoothness & 98.64 & 98.31 & 98.32 & 98.40 \\
Dynamic degree & 48.89 & 50.00 & 50.56 & 46.11 \\
Aesthetic quality & 67.02 & 67.26 & 67.46 & 67.27 \\
Imaging quality & 70.64 & 71.39 & 71.33 & 71.45 \\
Object class & 94.70 & 93.51 & 95.08 & 94.86 \\
Multiple objects & 80.20 & 84.68 & 83.17 & 81.71 \\
Human action & 99.40 & 98.80 & 99.00 & 98.60 \\
Color & 83.80 & 83.11 & 83.35 & 83.32 \\
Spatial relationship & 81.61 & 85.03 & 85.84 & 85.53 \\
Scene & 53.58 & 54.51 & 55.77 & 54.55 \\
Appearance style & 21.30 & 21.36 & 21.21 & 21.19 \\
Temporal style & 25.28 & 25.51 & 25.48 & 25.47 \\
Overall consistency & 27.32 & 27.47 & 27.38 & 27.47 \\
\midrule
Quality aggregate$\uparrow$ & 83.56 & 83.73 & 83.71 & 83.46 \\
Semantic aggregate$\uparrow$ & 80.43 & 81.30 & 81.52 & 81.11 \\
Total$\uparrow$ & 82.94 & 83.24 & 83.28 & 82.99 \\
\bottomrule
\end{tabular}
\end{table*}

\subsection{Dimension-level quality}
Most of the 80\%-to-90\% local-query decline comes from Dynamic Degree: its raw score decreases from 50.56 to 46.11 and contributes approximately 0.27 of the 0.29 total-point decline after official normalization. Dynamic Degree measures the amount of motion, not motion correctness.

On VBench2.0, HetA improves over dense DMD3 in Instance Preservation (+5.26), Motion Order Understanding (+5.05), Complex Landscape (+4.00), Material (+2.22) and Dynamic Spatial Relationship (+0.48), and matches it in Camera Motion.

\begin{figure*}[t]
\centering
\begin{minipage}{0.48\linewidth}
\includegraphics[width=\linewidth]{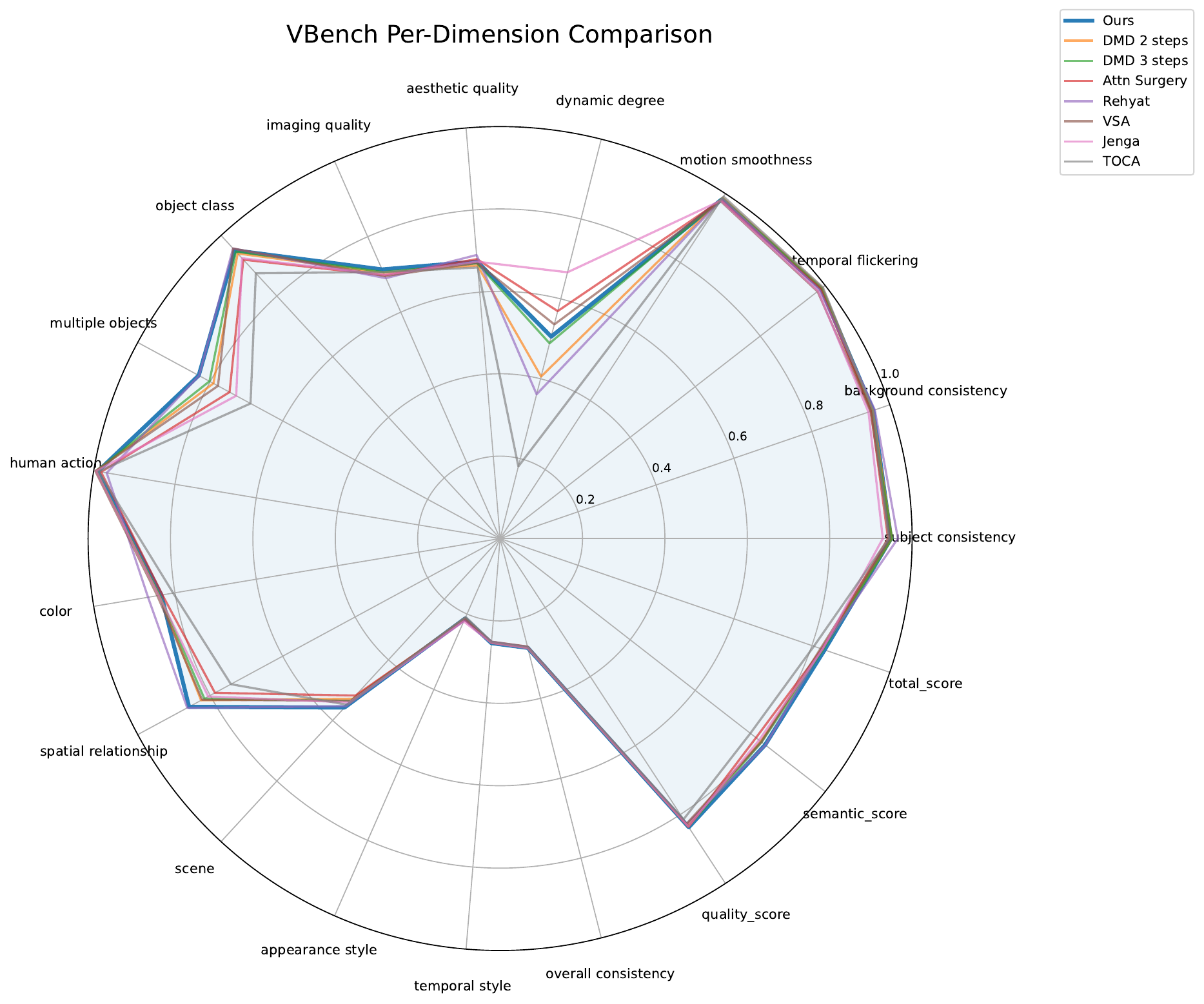}
\end{minipage}\hfill
\begin{minipage}{0.48\linewidth}
\includegraphics[width=\linewidth]{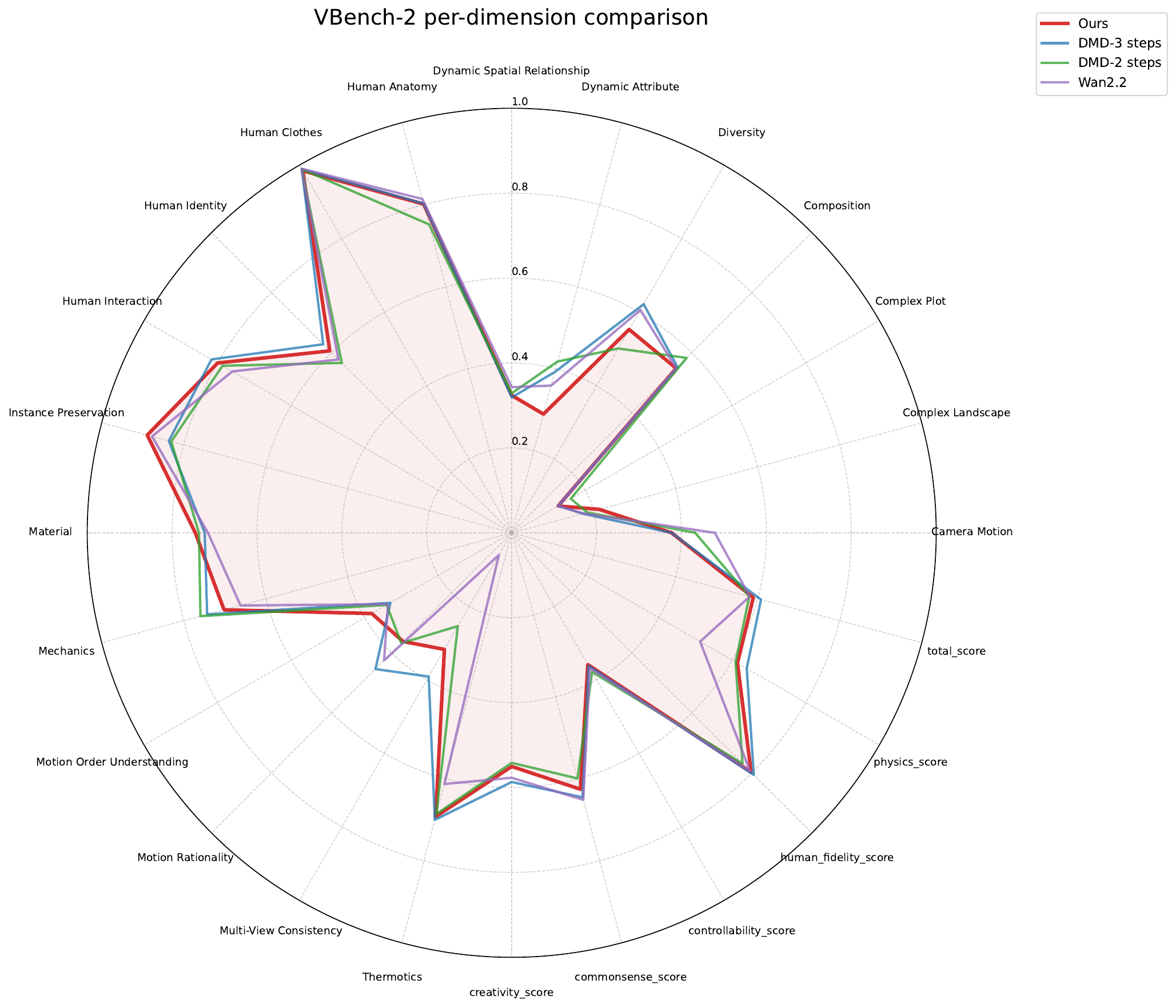}
\end{minipage}
\caption{Per-dimension VBench (left) and VBench2.0 (right) profiles.}
\label{fig:supp-vbench}
\end{figure*}

\subsection{Random swaps}
Swaps expose tokens outside the uncertainty-selected set to global context while preserving the global-query count. If a block promotes $m$ queries drawn uniformly from a base local set of $N_L$ queries, a member of that set is promoted at least once in $B$ independent blocks with probability $1-(1-m/N_L)^B$. For $N_L=0.8N$, $m=0.07N$ and $B=30$, this is approximately 0.94. The conditional probability is $m/N_L$, since the swap fraction is defined relative to all tokens.

\subsection{Wan2.1-14B qualitative examples}
The following comparisons use the same evaluated model as Table~\ref{tab:wan14_quality}, with 3,000 attention-adaptation generator updates and training noise shift 5. All examples use the first VBench sample for each prompt, without matching inference seeds across methods.
\begin{figure}[p]
\centering
\includegraphics[width=\linewidth]{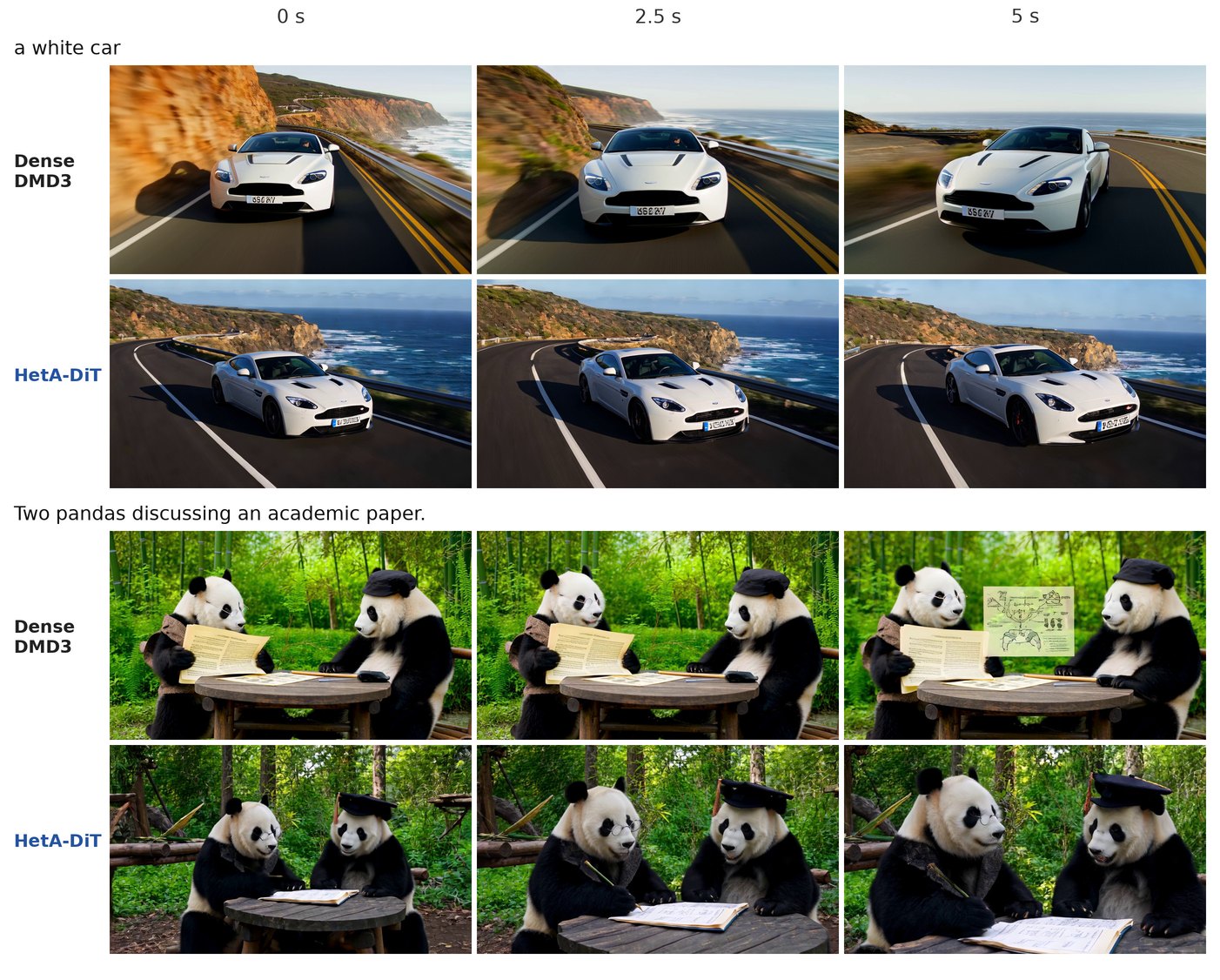}
\caption{\textbf{Wan2.1-14B at three denoising steps.} Dense DMD3 (upper row per prompt) and the evaluated HetA model (lower row), at 0, 2.5 and 5 seconds. Samples use the first VBench sample for the same prompts; inference seeds are not paired between methods.}
\label{fig:wan14_qualitative}
\end{figure}

\begin{figure}[p]
\centering
\includegraphics[width=\linewidth]{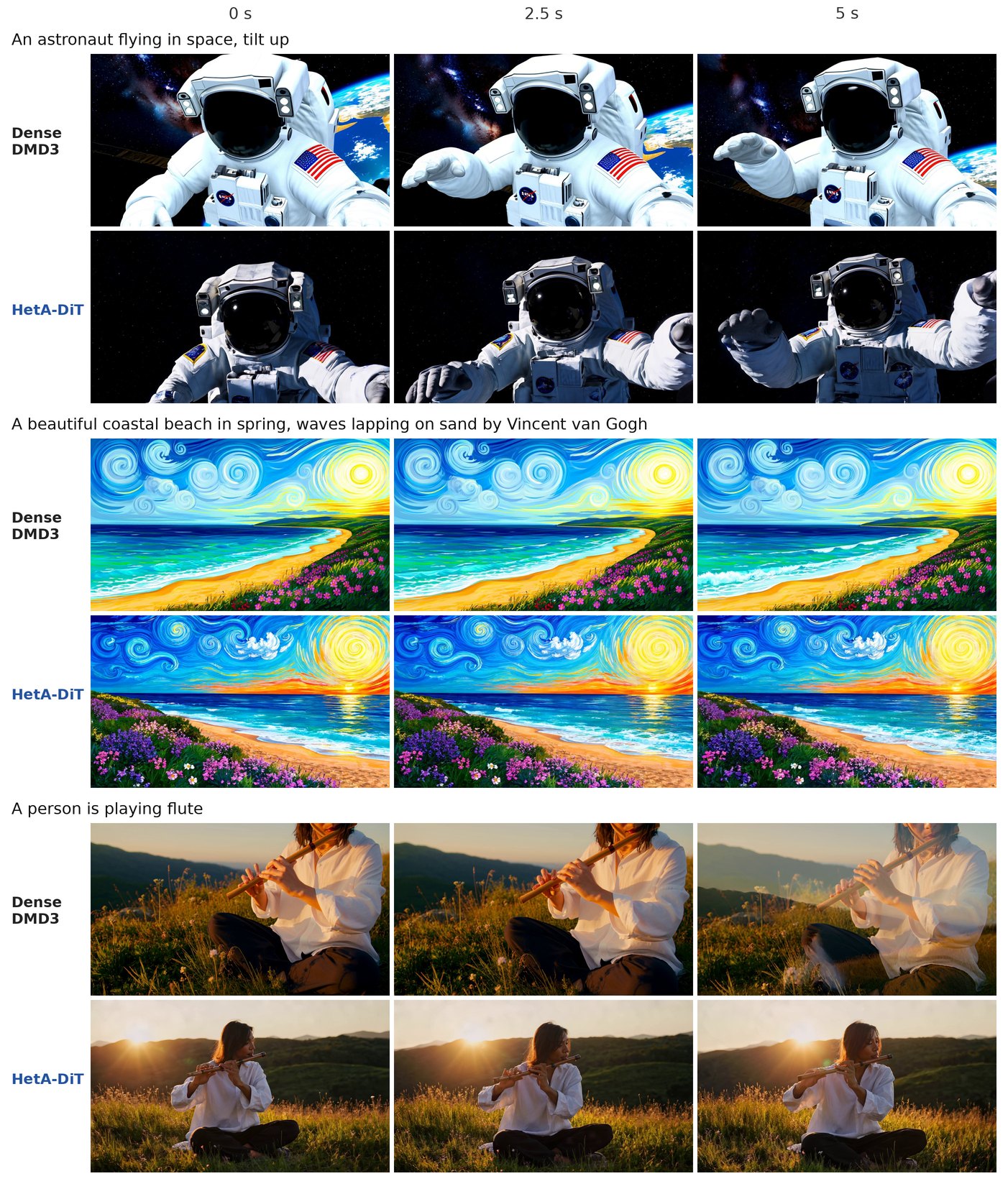}
\caption{Additional prompt-matched Wan2.1-14B examples: dense DMD3 above HetA-DiT for each prompt, at 0, 2.5 and 5 seconds. Both methods use three denoising steps and the original decoder.}
\label{fig:wan14_more_qualitative}
\end{figure}

\clearpage
\section{Additional Qualitative Comparisons}
\begin{figure}[htbp]
    \centering
    \includegraphics[width=1.0\linewidth]{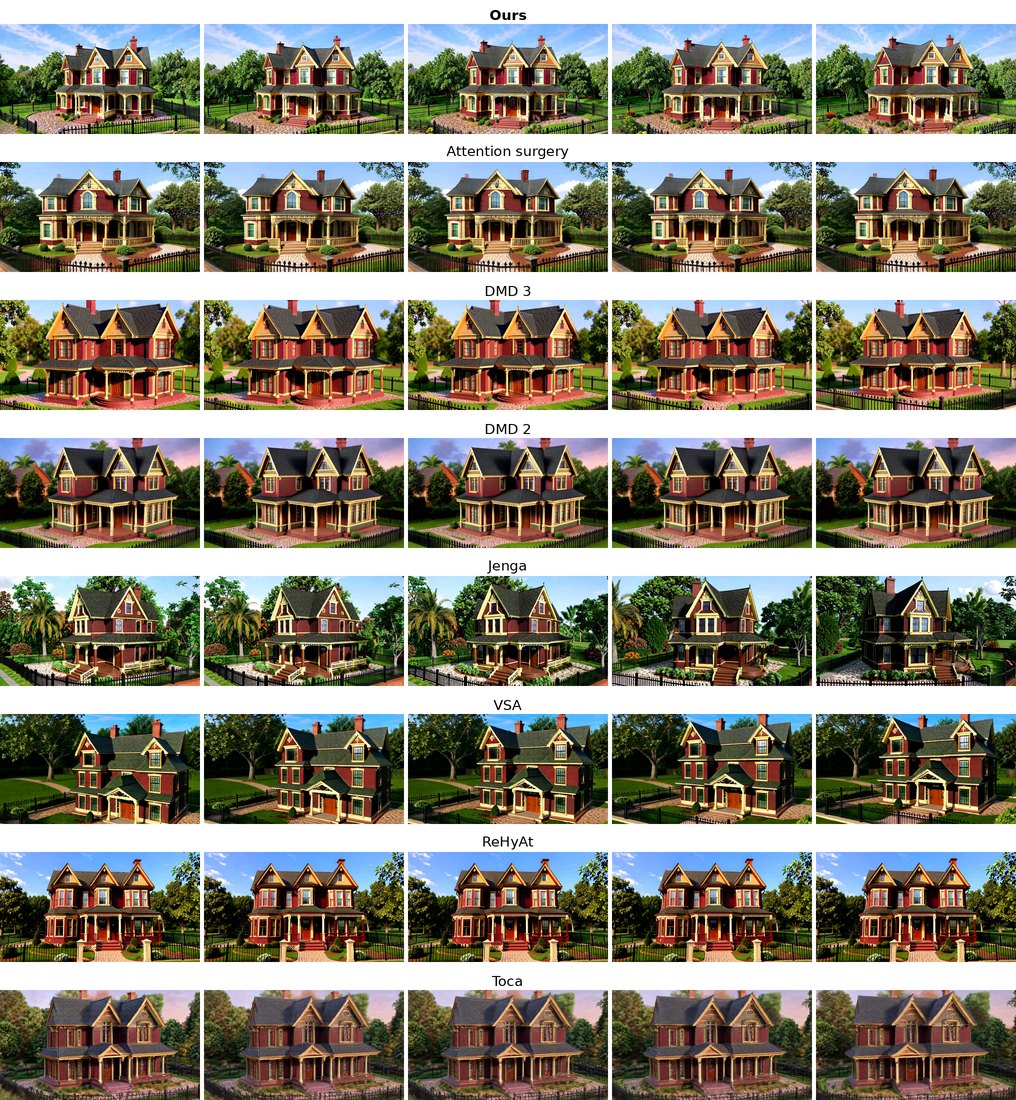}
    \caption{Qualitative comparison of HetA-DiT to the state-of-the-art efficient video generation models. \textbf{Prompt}: \textit{A 3D model of a 1800s victorian house.} }
    \label{fig:supp-qual-1}
\end{figure}

\begin{figure}[htbp]
    \centering
    \includegraphics[width=1.0\linewidth]{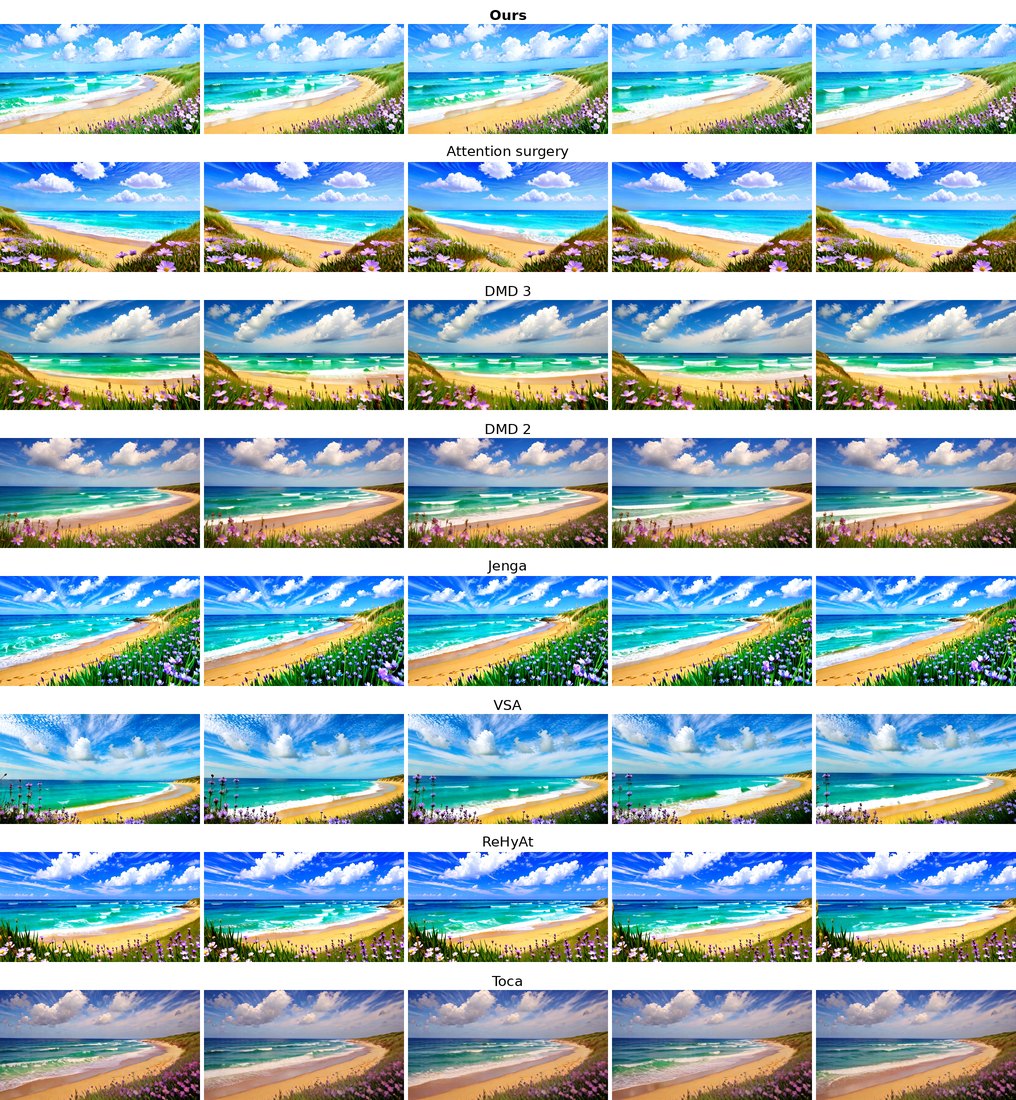}
    \caption{Qualitative comparison of HetA-DiT to the state-of-the-art efficient video generation models. \textbf{Prompt}: \textit{A beautiful coastal beach in spring, waves lapping on sand, Van Gogh style.}}
    \label{fig:supp-qual-2}
\end{figure}

\begin{figure}[htbp]
    \centering
    \includegraphics[width=1.0\linewidth]{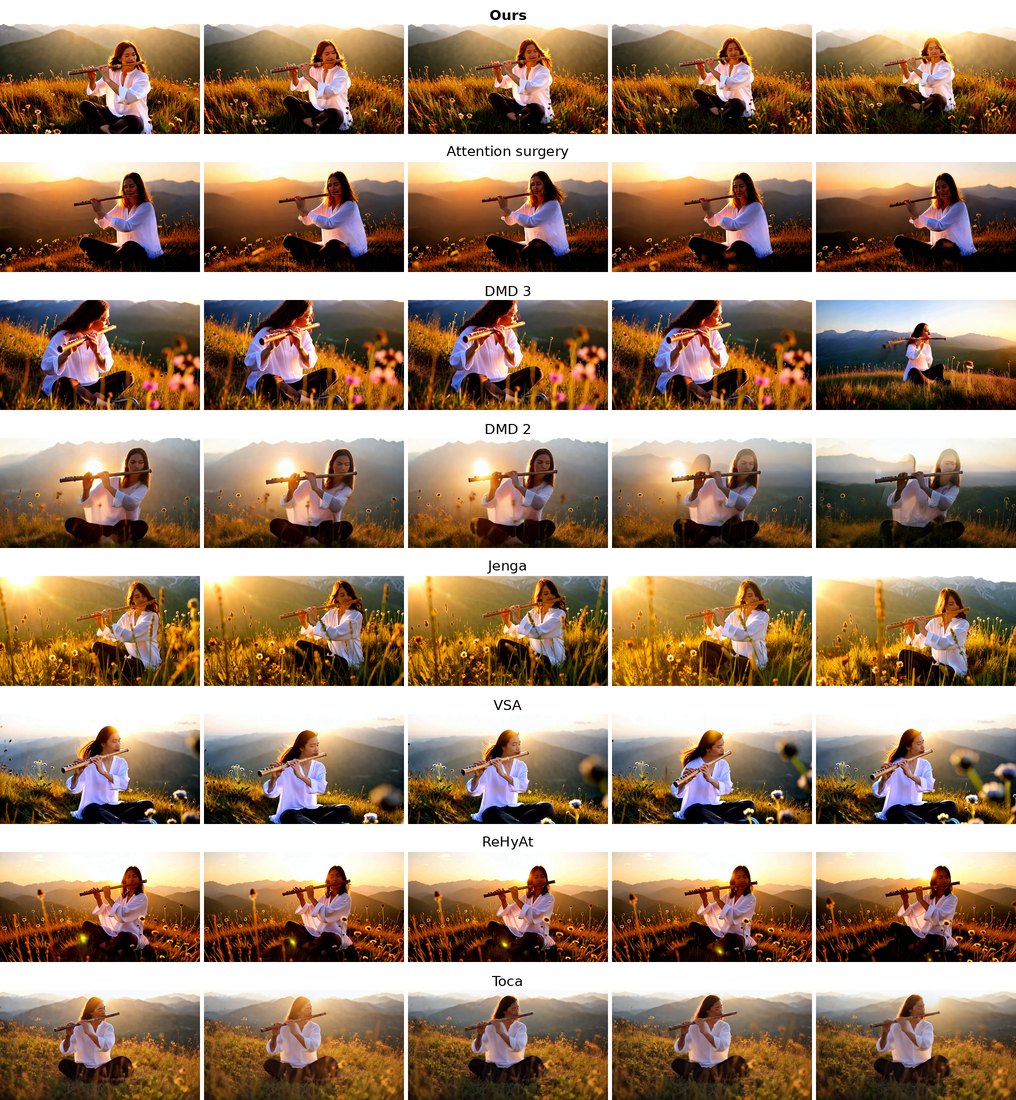}
    \caption{Qualitative comparison of HetA-DiT to the state-of-the-art efficient video generation models. \textbf{Prompt}: \textit{A person is playing flute.}}
    \label{fig:supp-qual-3}
\end{figure}

\begin{figure}[htbp]
    \centering
    \includegraphics[width=1.0\linewidth]{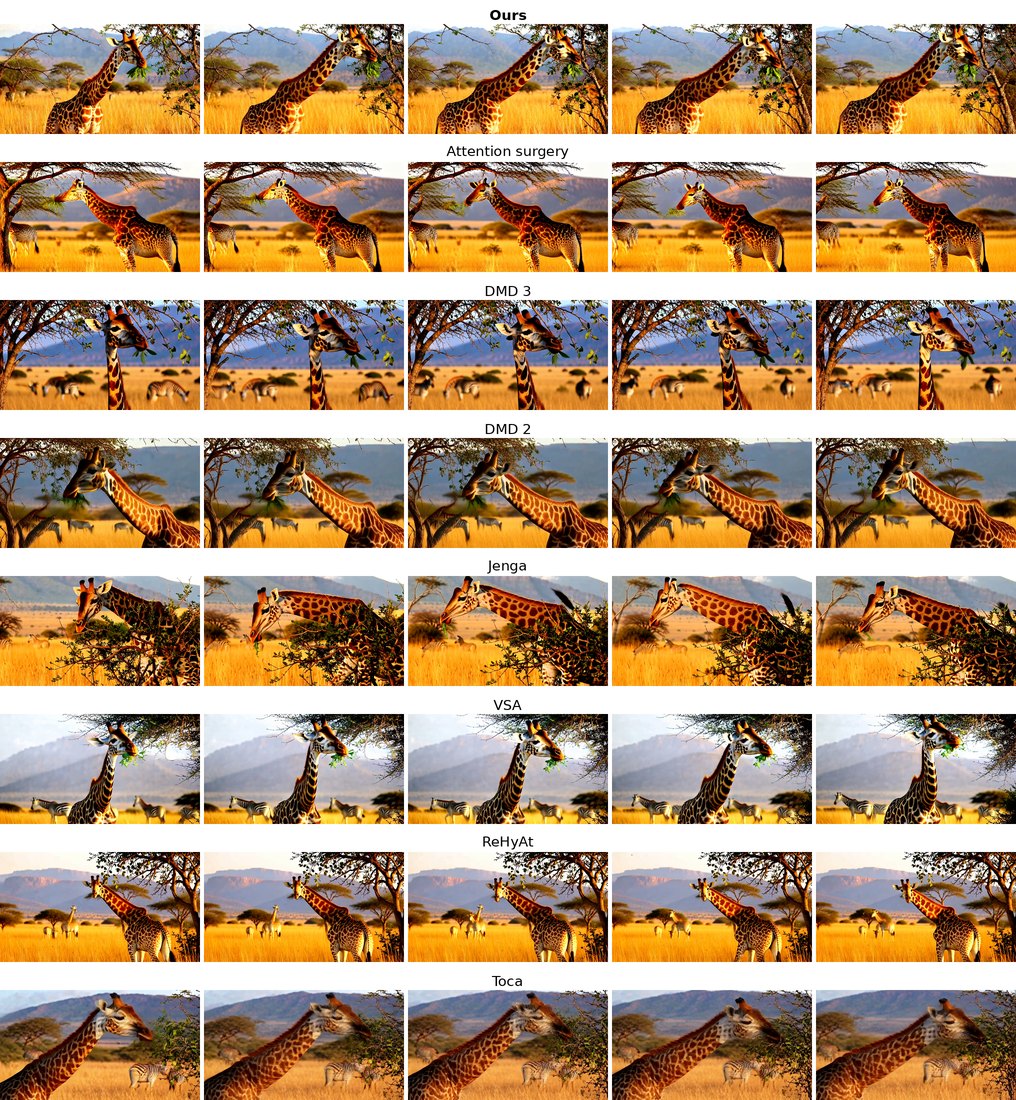}
    \caption{Qualitative comparison of HetA-DiT to the state-of-the-art efficient video generation models. \textbf{Prompt}: \textit{a giraffe}.}
    \label{fig:supp-qual-4}
\end{figure}

\begin{figure}[htbp]
    \centering
    \includegraphics[width=1.0\linewidth]{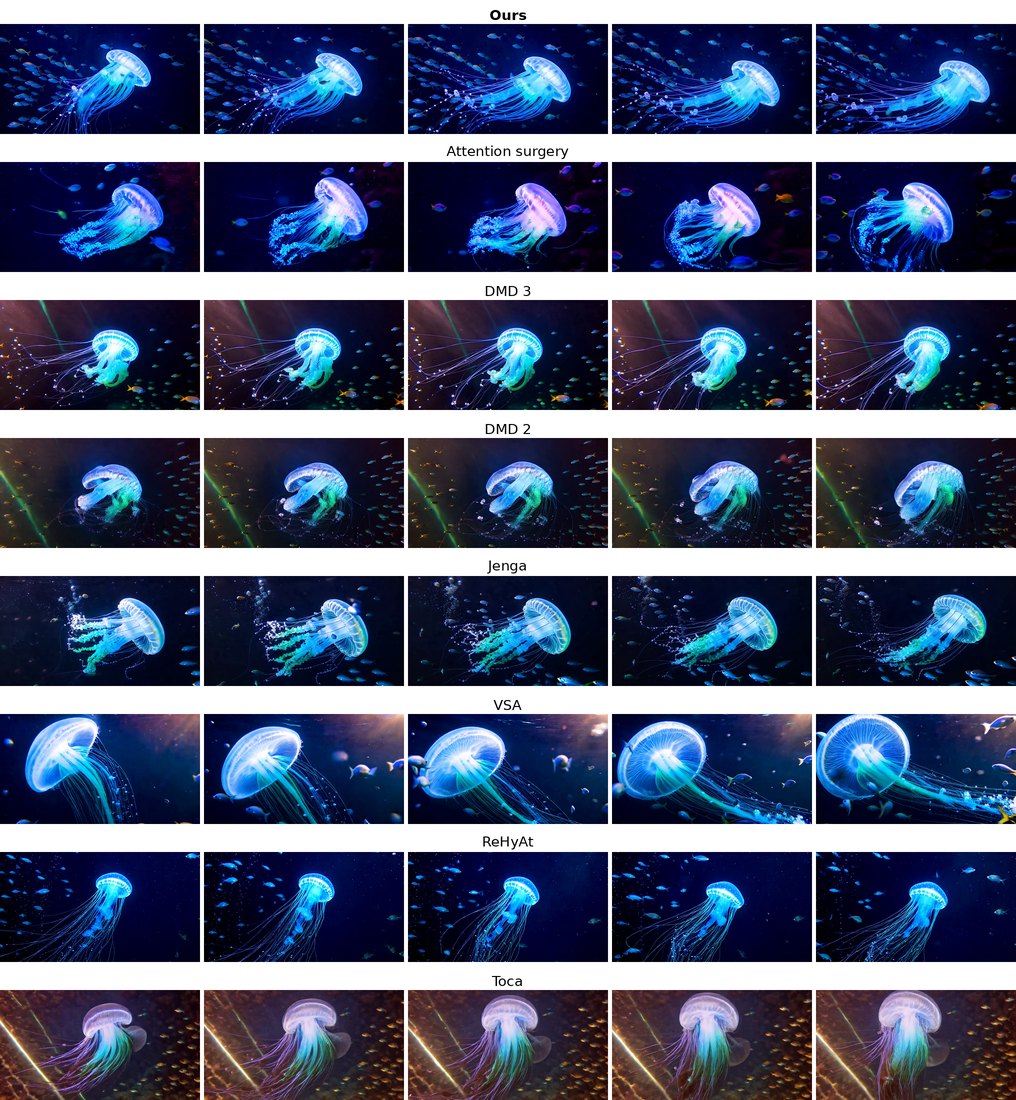}
    \caption{Qualitative comparison of HetA-DiT to the state-of-the-art efficient video generation models. \textbf{Prompt}: \textit{A jellyfish floating through the ocean, with bioluminescent tentacles}.}
    \label{fig:supp-qual-5}
\end{figure}

We provide additional qualitative comparisons of our methods to other efficient video generation baselines in Figure~\ref{fig:supp-qual-1}, Figure~\ref{fig:supp-qual-2}, Figure~\ref{fig:supp-qual-3}, Figure~\ref{fig:supp-qual-4} and Figure~\ref{fig:supp-qual-5}. 

\clearpage
\section{Uncertainty Validation and Interpretation}
\label{sec:uncertainty_validation}
\subsection{Whole-denoiser global-context sensitivity}
For each later denoising step, we evaluate the complete trained 30-block Wan2.2-5B denoiser twice, using dense attention or local attention everywhere except the mandatory global first latent frame. We hold weights, noisy latent, text and timestep fixed. For token $j$, discrepancy $d_j$ is the RMS difference of the predicted denoising updates over its output patch (48 latent channels and four spatial values per channel). We compare $d_j$ with uncertainty predicted one denoising step earlier, which is available when HetA chooses its route.

Spearman correlation compares the two rankings. We calculate one correlation per input and then average across inputs. Each cohort has eight prompts, two seeds and both later steps, totaling 32 inputs. The second cohort uses eight disjoint prompts fixed before collection. Confidence intervals use 20,000 prompt-cluster resamples, keeping each prompt's seeds and steps together.

\begin{table*}[t]
\centering\small
\setlength{\tabcolsep}{8pt}
\begin{tabular}{lcc}
\toprule
Quantity & Initial eight prompts & Eight disjoint prompts \\
\midrule
Mean Spearman correlation & 0.319 [0.227, 0.415] & 0.359 [0.289, 0.439] \\
Selected eligible tokens & 17.40\% & 18.25\% \\
Squared discrepancy captured & 21.94\% & 23.60\% \\
Capture / random expectation & $1.26\times$ [1.20, 1.33] & $1.29\times$ [1.21, 1.38] \\
Training-target correlation & 0.522 & 0.545 \\
\bottomrule
\end{tabular}
\caption{Previous-step uncertainty versus dense/local whole-denoiser output discrepancy. Brackets give 95\% prompt-cluster intervals. The final row is a separate check on learning the normalized DMD-residual training target.}
\label{tab:uncertainty_diagnostic}
\end{table*}

\begin{figure*}[t]
\centering
\includegraphics[width=\linewidth]{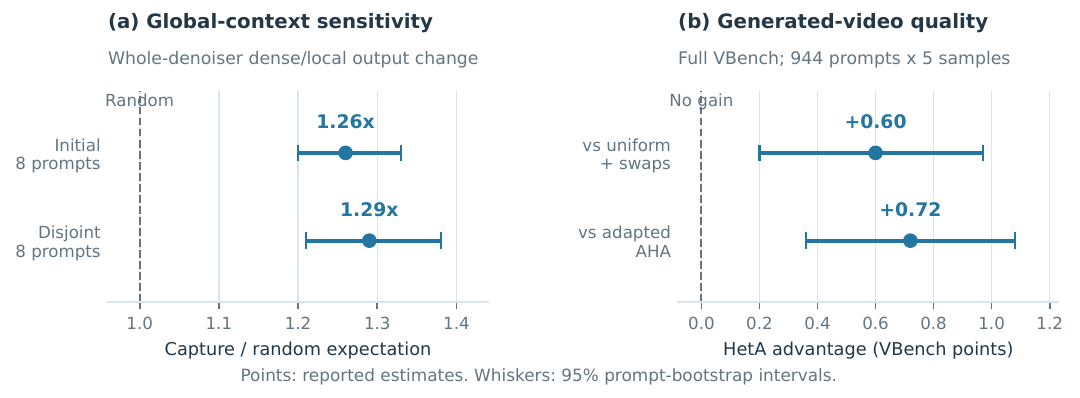}
\caption{\textbf{Uncertainty identifies sensitivity, and guided routing improves quality.} (a) Squared output-discrepancy capture relative to random selection in the initial and disjoint diagnostic cohorts. (b) Full-VBench gains over uniform routing with matched swaps and adapted AHA. Whiskers are 95\% prompt-bootstrap intervals; dashed lines mark random capture and zero quality gain. The panels use distinct prompt sets.}
\label{fig:routing_evidence}
\end{figure*}

\begin{figure*}[t]
\centering
\includegraphics[width=\linewidth]{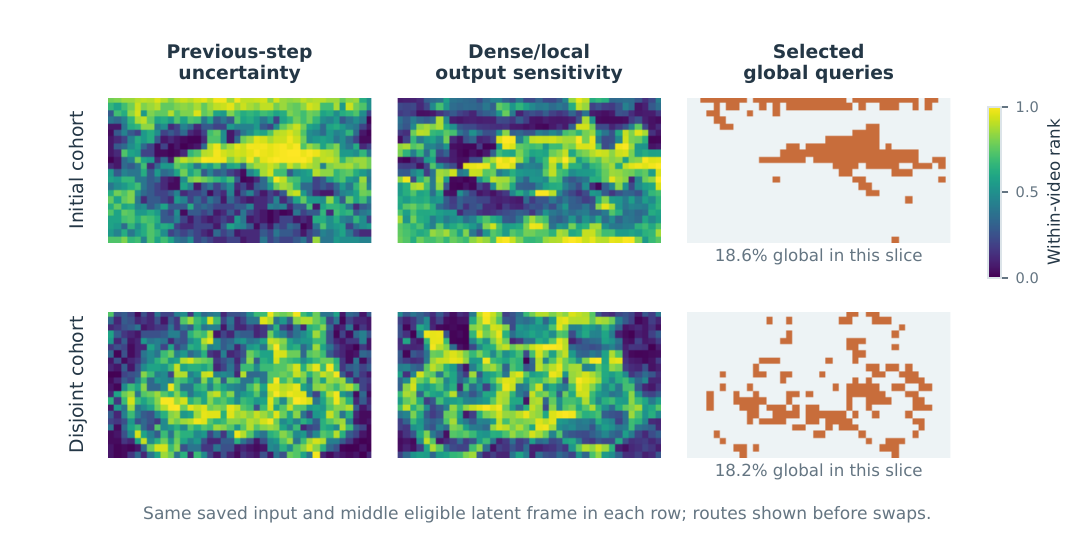}
\caption{\textbf{Spatial uncertainty, global-context sensitivity and selected queries.} Each row uses the first prompt, first seed and first later denoising step of its cohort, displaying the middle eligible latent frame. The first two columns show within-video rank percentiles of preceding-step uncertainty and dense/local whole-denoiser output discrepancy. Orange cells show the saved global-query selection before swaps. Always-global first-frame tokens are excluded. These examples illustrate spatial structure; the aggregate correlations and capture intervals above quantify the relationship across prompts.}
\label{fig:uncertainty_spatial}
\end{figure*}

Capture uses threshold-selected queries before swaps, excluding always-global first-frame tokens. Hence the nominal 20\% global budget gives different observed eligible fractions. Uniformly selecting a fraction $p$ captures $p$ of squared discrepancy in expectation. We compute observed capture and enrichment per input and then average. The disjoint cohort's 18.25\% selection captures 23.60\%, or $1.29\times$ random expectation. This measures output sensitivity to global context.

\subsection{Training-target check}
Uncertainty correlates with its supervised normalized DMD-residual target at 0.522 and 0.545 in the two cohorts. This confirms that the branch learns its training target.

\subsection{SNR interpretation}

We provide intuition behind the proposed uncertainty analysis. More specifically, the uncertainty prediction can be motivated through a simple SNR-based interpretation. Suppose that, for a given token, the model prediction can be viewed as a Gaussian estimate of the clean latent:
\begin{equation}
x_0 \approx \hat{x}_0 + u\eta,
\qquad
\eta \sim \mathcal{N}(0,1).
\label{eq:uncertainty_gaussian}
\end{equation}
Equivalently, $\hat{x}_0 \approx x_0 - u\eta$. Re-noising this prediction to timestep $\tau$ gives
\begin{equation}
\begin{aligned}
\hat{x}_\tau
&=
\alpha_\tau \hat{x}_0 + \sigma_\tau \epsilon \\
&=
\alpha_\tau x_0 + \sigma_\tau \epsilon - \alpha_\tau u\eta .
\end{aligned}
\label{eq:renoising}
\end{equation}
Assuming that $\epsilon$ and $\eta$ are independent standard Gaussian variables, the effective noise variance becomes
\begin{equation}
\sigma_\tau^2 + \alpha_\tau^2 u^2.
\end{equation}
Thus, the effective signal-to-noise ratio is
\begin{equation}
\mathrm{SNR}_{\mathrm{eff}}(\tau,u)
=
\frac{\alpha_\tau^2}
{\sigma_\tau^2 + \alpha_\tau^2 u^2}.
\label{eq:effective_snr}
\end{equation}
We match this quantity to the SNR of an artificial timestep $\tau'$:
\begin{equation}
\frac{\alpha_\tau^2}
{\sigma_\tau^2 + \alpha_\tau^2 u^2}
=
\frac{\alpha_{\tau'}^2}{\sigma_{\tau'}^2},
\qquad
\tau' \geq \tau.
\label{eq:snr_matching}
\end{equation}
Under this interpretation, a larger uncertainty $u$ makes the token behave as if it were at a noisier timestep $\tau'$. This motivates using uncertainty as a routing signal; the dense/local sensitivity study above tests its empirical relationship with global context. This is similar in spirit to SNR matching in pyramidal or multi-scale diffusion~\citep{jin2024pyramidalflow}.

\begin{figure*}
    \centering
    \includegraphics[width=1.0\linewidth]{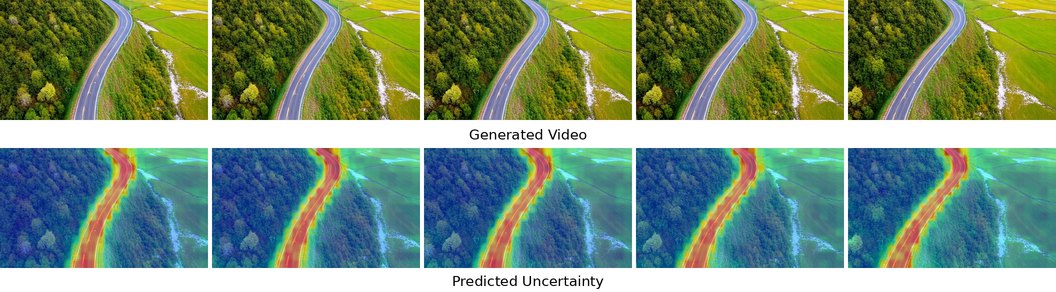}
    \caption{Example of the predicted uncertainty map at diffusion timestep $\sigma=0.6$.}
    \label{fig:supp-unc-1}
\end{figure*}

\begin{figure*}
    \centering
    \includegraphics[width=1.0\linewidth]{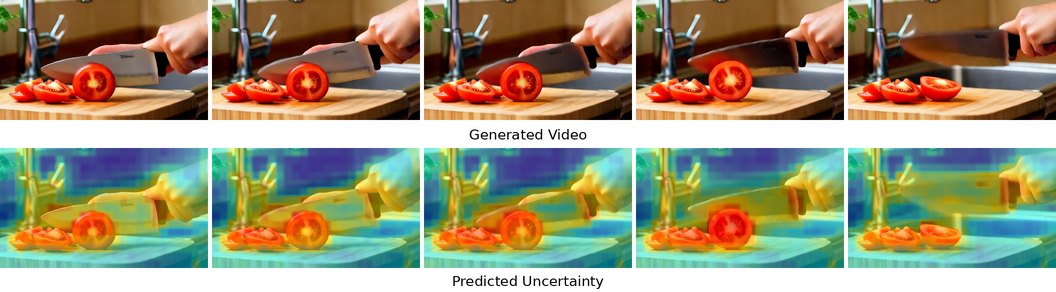}
    \caption{Example of the predicted uncertainty map at diffusion timestep $\sigma=0.6$.}
    \label{fig:supp-unc-2}
\end{figure*}

\begin{figure*}
    \centering
    \includegraphics[width=1.0\linewidth]{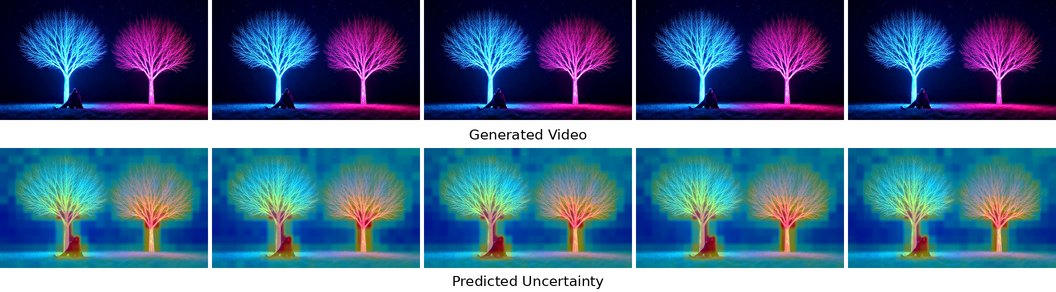}
    \caption{Example of the predicted uncertainty map at diffusion timestep $\sigma=0.6$.}
    \label{fig:supp-unc-3}
\end{figure*}

\begin{figure*}
    \centering
    \includegraphics[width=1.0\linewidth]{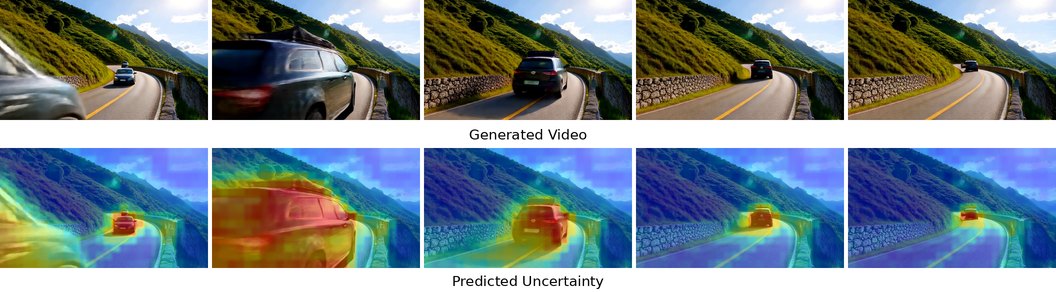}
    \caption{Example of the predicted uncertainty map at diffusion timestep $\sigma=0.6$.}
    \label{fig:supp-unc-4}
\end{figure*}

\begin{figure*}
    \centering
    \includegraphics[width=1.0\linewidth]{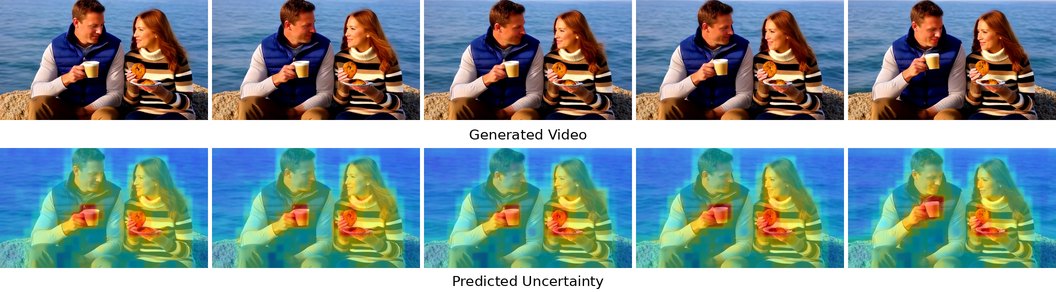}
    \caption{Example of the predicted uncertainty map at diffusion timestep $\sigma=0.6$.}
    \label{fig:supp-unc-5}
\end{figure*}

We further show some qualitative examples of the predicted uncertainty maps in Figure~\ref{fig:supp-unc-1}, Figure~\ref{fig:supp-unc-2}, Figure~\ref{fig:supp-unc-3}, Figure~\ref{fig:supp-unc-4} and Figure~\ref{fig:supp-unc-5}. 

\end{document}